\documentclass[10pt,twocolumn,letterpaper]{article}

\usepackage[pagenumbers]{cvpr} 

\usepackage{graphicx}
\usepackage{amsmath}
\usepackage{amssymb}
\usepackage{booktabs}
\usepackage{makecell}
\makeatletter
\@namedef{ver@everyshi.sty}{}
\makeatother
\usepackage{tikz}
\usetikzlibrary{tikzmark}

\usepackage[pagebackref,breaklinks,colorlinks]{hyperref}

\usepackage[capitalize]{cleveref}
\crefname{section}{Sec.}{Secs.}
\Crefname{section}{Section}{Sections}
\Crefname{table}{Table}{Tables}
\crefname{table}{Tab.}{Tabs.}

\def\cvprPaperID{7228} 
\def\confName{CVPR}
\def\confYear{2023}

\usepackage{xcolor}         
\usepackage{subfiles}
\usepackage{bm}
\usepackage{cite}
\usepackage{multirow}
\usepackage[accsupp]{axessibility} 
\usepackage{graphicx}
\usepackage{wrapfig}
\usepackage[font=small,labelfont=bf]{caption}
\usepackage{amsmath}
\usepackage{pifont}
\usepackage{colortbl}
\usepackage{enumitem}
\usepackage{multirow}
\usepackage{colortbl}
\usepackage{axessibility}
\setlist{nolistsep}

\newcommand{\Paragraph}[1]{\vspace{1mm}\noindent\textbf{#1.}\hspace{0mm}}
\newcommand{\Section}[1]{\vspace{-0mm} \section{#1} \vspace{-0mm}}

\begin{document}

\newcommand{\ethnicityratio}{

\begin{table}[t]
\centering
\scriptsize
\begin{tabular}{|l|l|l|l|l|l|}
\hline
 & White & Asian & Others & Black & Indian \\ \hline
CASIA-WebFace &
  \cellcolor[HTML]{FFC702}$0.634$ &
  \cellcolor[HTML]{FCFF2F}$0.144$ &
  \cellcolor[HTML]{FFFFC7}$0.074$ &
  \cellcolor[HTML]{FFFFC7}$0.074$ &
  \cellcolor[HTML]{FFFFC7}$0.072$ \\ \hline
 DDPM $G_{id}$ &
  \cellcolor[HTML]{FFC702}$0.660$ &
  \cellcolor[HTML]{FCFF2F}$0.209$ &
  \cellcolor[HTML]{FFFFC7}$0.034$ &
  \cellcolor[HTML]{FFFFC7}$0.046$ &
  \cellcolor[HTML]{FFFFC7}$0.048$ \\ \hline
Balanced Ethnicity &
  \cellcolor[HTML]{FFCE93}$0.200$ &
  \cellcolor[HTML]{FFCE93}$0.200$ &
  \cellcolor[HTML]{FFCE93}$0.200$ &
  \cellcolor[HTML]{FFCE93}$0.200$ &
  \cellcolor[HTML]{FFCE93}$0.200$ \\ \hline
\end{tabular}
\vspace{-2mm}
\caption{Ethnicity Distribution of CASIA-WebFace. Ethnicity prediction is made using~\cite{faceparsing}. DDPM $G_{id}$  is trained on FFHQ~\cite{karras2019style}. }
\label{tab:distribution}
\vspace{-2mm}
\end{table}

}

\definecolor{ttt}{rgb}{0.9, 0.9, 0.9}
\newcommand{\fillc}{\cellcolor{ttt}}
\definecolor{tttt}{rgb}{0.87, 0.87, 0.87}
\newcommand{\fillch}{\cellcolor{tttt}}

\newcommand{\modelAblation}{
\begin{table}[t]
\centering
\scriptsize
\setlength{\tabcolsep}{3pt}
\begin{tabular}{|c|c|c|c|c|c|c|}
    \hline
        Grid Size & Loss & Loss Model & $U_{class}$ & $C_{intra}$ & $D_{intra}$ & FR Perf. \\ \hline\hline
        SynFace & - & - & $0.080$ & $0.9966$ & $0.131$ & $74.75$ \\ 
        DigiFace & - & - & $0.178$ & $0.9973$ & $0.297$ & $83.45$ \\ \hline \hline
        \fillc $1\!\times\!1$ & \multirow{4}{*}{$L_{ID}$} & \multirow{4}{*}{$F$} & $\bm{0.978}$ & $\bm{0.9987}$ & $0.4418$ & $79.28$ \\ 
        \fillc $3\!\times\!3$ & ~ & ~ & $0.956$ & $0.9809$ & $0.7030$ & $85.79$ \\ 
        \fillc \textcolor{blue}{$\bm{5\!\times\!5}$} & ~ & ~ & $0.924$ & $0.9035$ & $0.7734$ & $\bm{89.04}$ \\ 
        \fillc $7\!\times\!7$ & ~ & ~ & $0.690$ & $0.5937$ & $\bm{0.7950}$ & $50.00$ \\ \hline\hline
        \multirow{3}{*}{$5\!\times\!5$} & \fillc $L_{naive1}$ & \multirow{3}{*}{$F$} & $\bm{0.988}$ & $\bm{0.9996}$ & $0.6546$ & $84.75$ \\ 
        ~ & \fillc $L_{naive2}$ & ~ & $0.866$ & $0.8046$ & $0.7835$ & $50.00$ \\ 
        ~ & \fillc \textcolor{blue}{$\bm{L_{ID}}$} & ~ & $0.924$ & $0.9035$ & $\bm{0.7734}$ & $\bm{89.04}$ \\ \hline\hline
        \multirow{2}{*}{$5\!\times\!5$} & \multirow{2}{*}{$L_{ID}$} & \fillc  \textcolor{blue}{$\bm{F}$} & $0.924$ & $0.9035$ & $0.7734$ & $89.04$ \\ 
        ~ & ~ & \fillc $F_{bigger}$ & $\bm{0.954}$ & $\bm{0.9197}$ & $0.7715$ & $\bm{89.89}$ \\ \hline
    \end{tabular}
    \vspace{-2mm}
\caption{Model Ablation. For FR performance, we generate a synthetic dataset of $10K$ subjects with $50$ images per subject using ($\textit{random}$, $\textit{random}$) ID and style sampling strategy. Blue color indicates the adopted setting for subsequent experiments. }
\label{table:modelAblation}
\vspace{-4mm}
\end{table}
}

\newcommand{\samplingAblation}{
\begin{table}[t]
\centering
\scriptsize
\setlength{\tabcolsep}{3pt}
    \begin{tabular}{|c|c|c|c|c|c|c||c|}
    \hline
        ID & Style & LFW & CFPFP & CPLFW & AGEDB & CALFW & AVG \\ \hline
        $\textit{random}$ & $\textit{random}$ & $98.05$  & $84.17$ &  $82.20$ &  $89.38$ &  $91.40$ &  $89.04$
 \\\hline
        $\textit{random}$ & $\textit{match}$ & $98.28$  & $84.61$ &  $82.32$ &  $89.12$ &  $91.28$ &  $89.12$
 \\\hline
        $\textit{balance}$ & $\textit{random}$ & $98.30$  & $83.27$ &  $81.60$ &  $89.40$ &  $91.27$ &  $88.77$
 \\\hline
        $\textit{balance}$ & $\textit{match}$ & $98.38$  & $84.06$ &  $82.45$ &  $89.30$ &  $91.38$ &  $89.11$
 \\\hline
        \textcolor{blue}{$\textit{balance}$} & \textcolor{blue}{$\textit{over smpl}$} & $\bm{98.55}$  &  $\bm{85.33}$ &  $\bm{82.62}$ &  $\bm{89.70}$ &  $\bm{91.60}$ &  $\bm{89.56}$ \\\hline
    \end{tabular}
\vspace{-2mm}
\caption{Sampling Ablation. We generate a synthetic dataset of $10K$ subjects with $50$ images per subject, using the setting indicated by the blue text in Tab.~\ref{table:modelAblation}. $\textit{over smpl}$ is over-sampling $\bm{X}_{id}$ during training for showing more front-view faces. }
\label{table:samplingAblation}
\vspace{-4mm}
\end{table}
}

\definecolor{gray}{rgb}{0.87, 0.87, 0.87}
\newcommand{\fillg}{\cellcolor{gray}}

\newcommand{\benchmarkTable}{
\begin{table*}[!ht]
    \centering
    \scriptsize
    \scalebox{1.08}{
    \begin{tabular}{|l|c|l|c|c|c|c|c||c||c|}
    \hline
        Methods & Venue  &\# images (\# IDs$\!\times\!$ \# imgs/ID) & LFW & CFP-FP & CPLFW & AgeDB & CALFW & Avg & \fillg Gap to Real \\ \hline\hline
        SynFace & ICCV21 & $0.5$M ($10$K $\!\times\!$ $50$)  & $91.93$ & $75.03$ & $70.43$ & $61.63$ & $74.73$ & $74.75$   & \fillg $26.58$ \\ 
        DigiFace & WACV23 & $0.5$M ($10$K $\!\times\!$ $50$) & $95.4$ & $\bm{87.4}$ & $78.87$ & $76.97$ & $78.62$ & $83.45$    & \fillg  $13.39$ \tikzmark{a}\\ 
        DCFace (Ours)     & - & $0.5$M ($10$K $\!\times\!$ $50$) &  $\bm{98.55}$  &  $ 85.33 $ &  $\bm{82.62}$ &  $\bm{89.70}$ &  $\bm{91.60}$ &  $\bm{89.56}$  & $\fillg  \bm{5.65}$ \tikzmark{b} \\\hline 
        DigiFace & WACV23  & $1.2$M ($10$K $\!\times\!$ $72$ + $100$K $\!\times\!$ 5) & $96.17$  & $\bm{89.81}$ &  $82.23$ &  $81.10$ &  $82.55$  & $86.37$   &  \fillg  $9.55$  \tikzmark{c} \\ 
        DCFace (Ours)     &- & $1.0$M ($20$K $\!\times\!$ $50$)  &   $\bm{98.83}$  & $88.4$  &  $84.22$ &  $90.45$ &  $92.38$ &  $90.86$     &  \fillg  $4.14$ \\ 
        DCFace (Ours)     &- & $1.2$M ($20$K $\!\times\!$ $50$ + $40$K $\!\times\!$ $5$)  &  $98.58$  & $88.61$ &  $\bm{85.07}$ &  $\bm{90.97}$  & $\bm{92.82}$  & $\bm{91.21}$  &\fillg   $\bm{3.74}$ \tikzmark{d} \\ \hline\hline
        \multicolumn{2}{|c|}{CASIA-WebFace (Real)}  & $0.49$M (approx. $10.5$K $\!\times\!$ $47$)  &  $99.42$ &  $96.56$ &  $89.73$  & $94.08$  & $93.32$ &  $94.62$   & \fillg  $0.0$ \\ \hline
    \end{tabular}
    }
    \caption{\small Verification accuracies of FR models trained with SoTA synthetic training datasets. SynFace~\cite{qiu2021synface} is a GAN-based dataset with a latent space mixup technique. DigiFace~\cite{bae2022digiface} is a 3D model-based dataset with heavy image augmentation. DCFace uses the model setting from the ablation study, Tab.~\ref{table:modelAblation},~\ref{table:samplingAblation} indicated by blue colors. FR backbone is IR-SE50~\cite{deng2019arcface} + AdaFace~\cite{kim2022adaface} to match the setting of DigiFace. \vspace{-1.5mm}}
    \label{tab:comparison}
\end{table*}
}

\newcommand{\figone}{
\begin{figure}[t]
    \centering
    \includegraphics[width=0.96\linewidth]{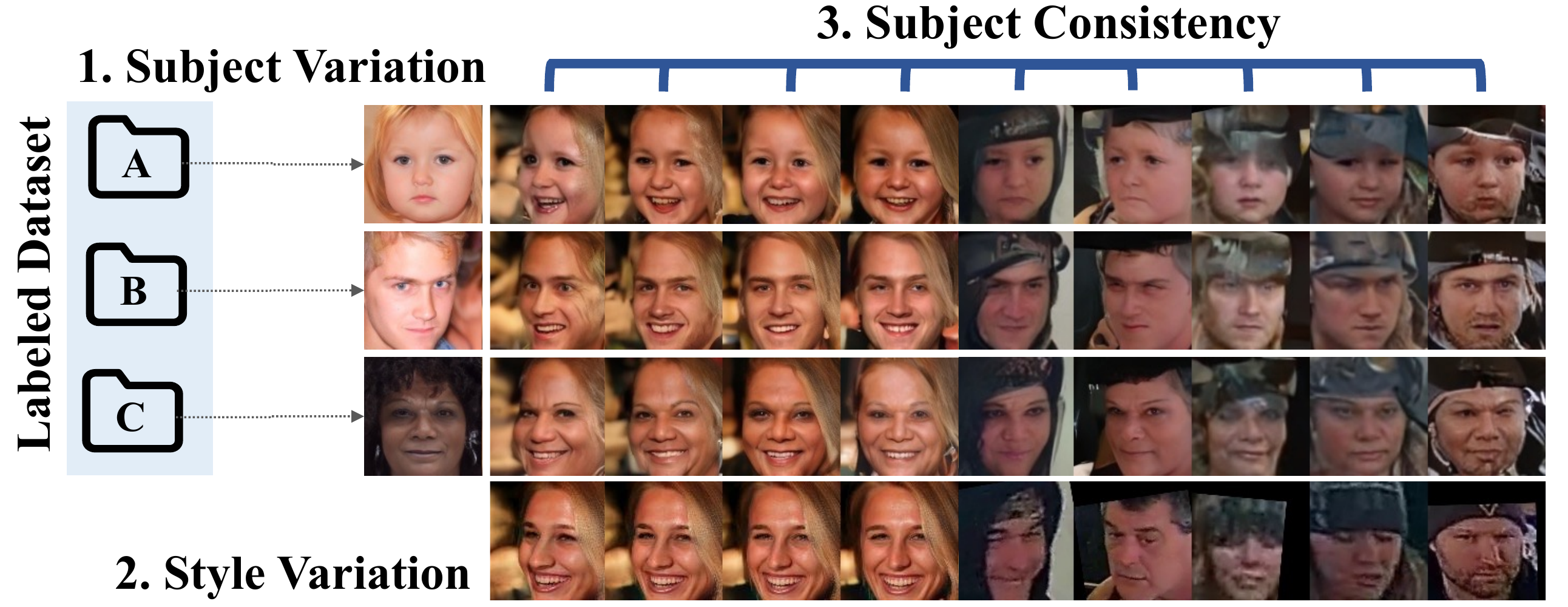}
    \caption{Illustration of three factors that characterize a labeled face dataset. It contains large subject variation, style variation and label consistency. Synthetic face datasets should be created with all three factors in mind. Face images in this figure are samples generated by our proposed method which combines arbitrary ID condition with style condition while preserving subject identity.}
    \label{fig:figure1}
    \vspace{-2mm}
\end{figure}
}

\newcommand{\figtwo}{
\begin{figure*}[t]
    \centering
    \includegraphics[width=\linewidth]{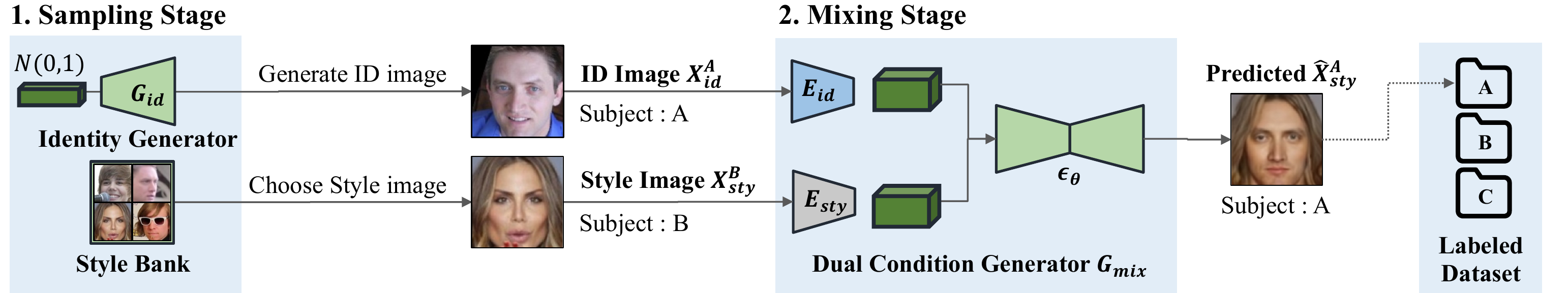}
    \caption{Two stage dataset generation paradigm. In the sampling stage, 1) $G_{id}$ generates a high-quality face image $\mathbf{X}_{id}$ that defines how a person looks and 2) the style bank selects a style image $\mathbf{X}_{sty}$ that defines the overall style of the final image. The mixing stage generates image with identity from $\mathbf{X}_{id}$ and style from $\mathbf{X}_{sty}$. Repeating this process multiple times, one can generate a labeled synthetic face dataset.  }
    \label{fig:figure2}
\end{figure*}
}

\newcommand{\figGANDDPM}{
\begin{figure}[t]
    \centering
    \includegraphics[width=0.82\linewidth]{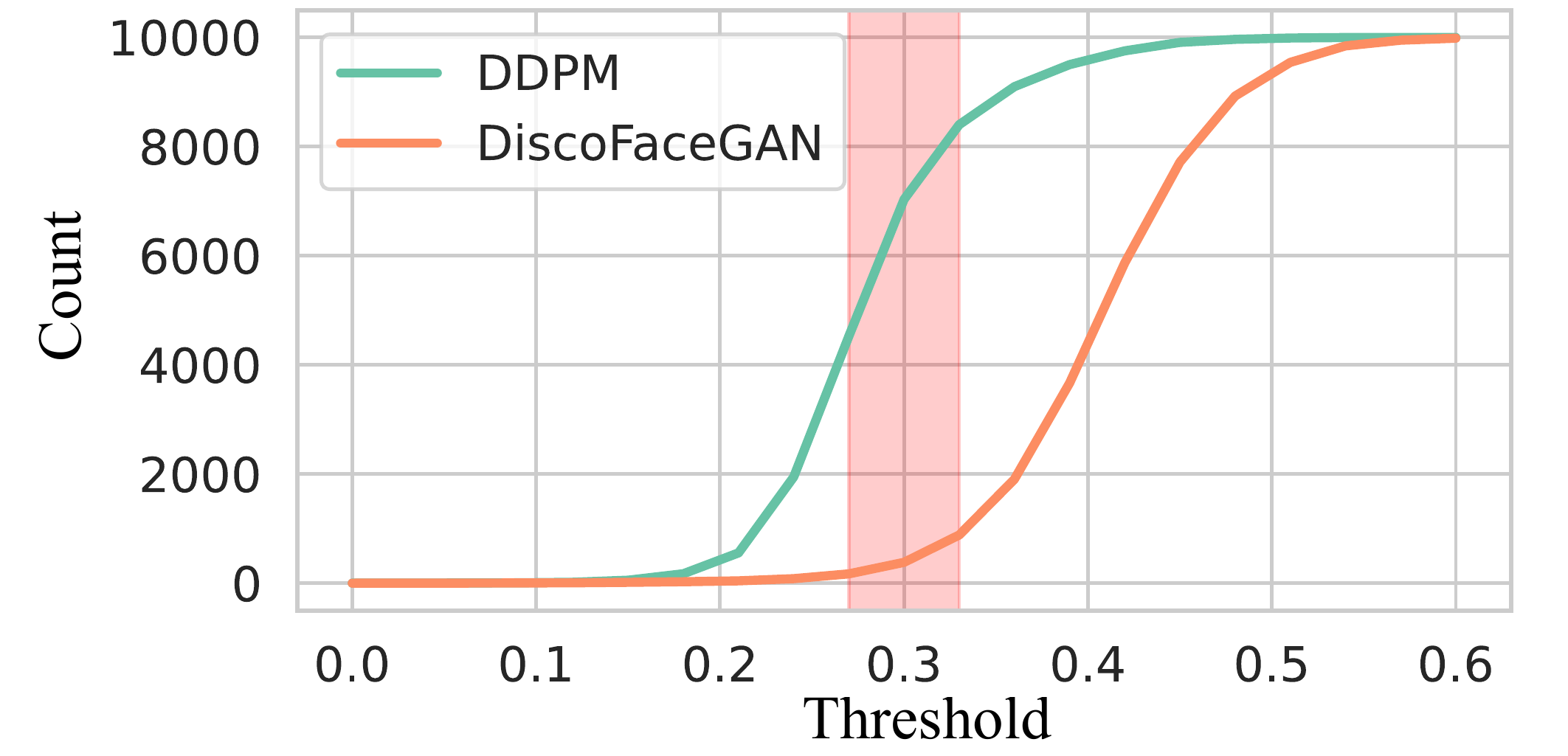}
    \vspace{-2mm}
    \caption{Comparison of the number of unique subjects generated by DiscoFaceGAN~\cite{deng2020disentangled} and unconditional DDPM~\cite{ho2020denoising}. Uniqueness is the number of unique subjects measured by a face recognition model. By varying the threshold which determines a match between two subjects, we plot the number of unique subjects as defined in Eq.~\ref{eq:unique}. Unconditional DDPM and DiscoFaceGAN are trained on FFHQ~\cite{karras2019style} and each  generates $10,000$ samples.  The ability to generate novel subjects is larger for DDPM. Refer to Supp.E for additional details on the threshold.}
    \label{fig:ganddpmcomp}
    \vspace{-2mm}
\end{figure}
}

\newcommand{\algorithm}{
\begin{figure*}[t]
    \centering
    \includegraphics[width=\linewidth]{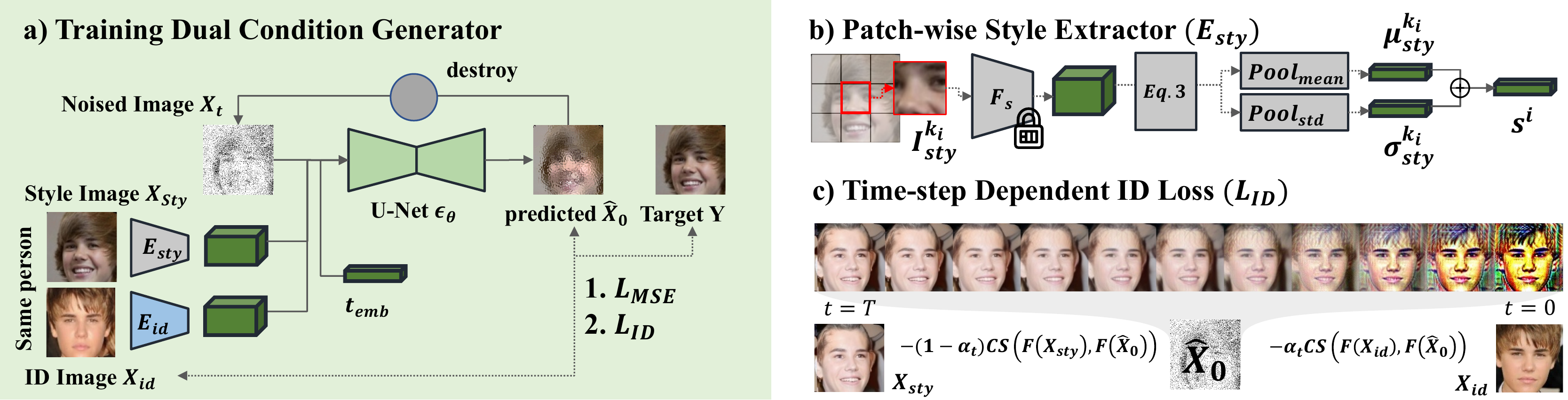}
    \caption{a) A diagram of $G_{mix}$ during training. At each step, we draw two labeled images from the labeled training dataset and use them as $\bm{X}_{id}$ and $\bm{X}_{sty}$. We ensure $\bm{X}_{id}$ to be the good-quality frontal view image. $t_{emb}$ is the time-step embedding in DDPM~\cite{ho2020denoising}. $\bm{X}_{sty}$ also serves as a target image and we apply Gaussian noise $\bm{\epsilon}$ to $\bm{X}_{sty}$ to create $\bm{X}_t$ as DDPM specifies. Then $\bm{\epsilon}_{\theta}(\bm{X}_{t},t,\bm{X}_{id}, \bm{X}_{sty})$ is trained to predict $\bm{\epsilon}$ using $L_{MSE}$, conceptually equivalent to the reconstruction loss to recover $\bm{X}_{sty}$. We also apply $L_{ID}$ as in Eq.~\ref{eq:lid} for the dependence on $\bm{X}_{id}$. b) Patch-wise Style Extractor generates style vectors from small patches of images. Style vectors are architecturally constrained from containing full ID information. c) Time-step dependent ID Loss is a linear interpolation between the $\bm{X}_{id}$ and $\bm{X}_{sty}$ in the recognition feature space. It forces $\bm{\epsilon}_\theta$ to rely on $\bm{X}_{id}$ to extract the subject's appearance and gradually shift the style to $\bm{X}_{sty}$. }
    \label{fig:algorithm}
\end{figure*}
}

\newcommand{\figfive}{
\begin{figure*}[t]
    \centering
    \includegraphics[width=\linewidth]{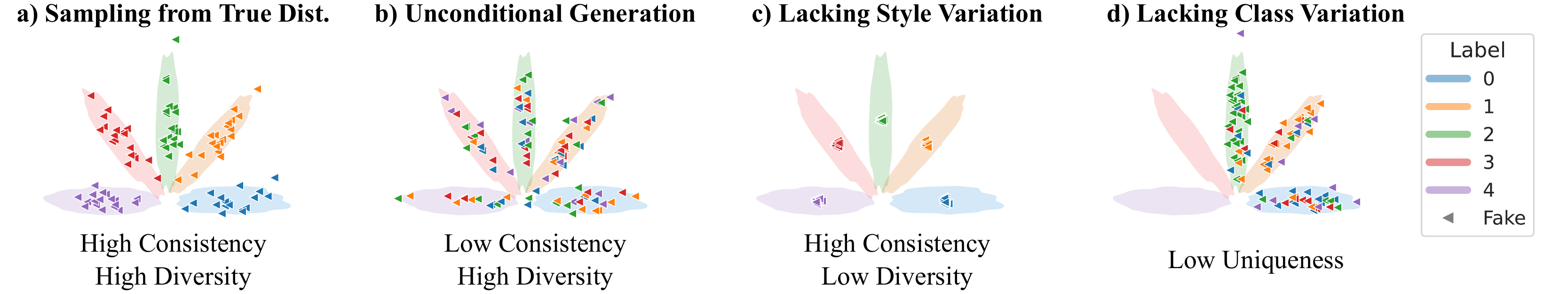}
    \vspace{-5mm}
    \caption{Illustration of conditional distributions in 2D space. Colored regions represent the true data distribution with individual colors representing different labels. Colored triangles represent generated samples with corresponding labels. For each scenario except (a), the generated distribution does not follow the true distribution. Consistency, diversity and uniqueness analysis can quantify the shortcomings. }
    \label{fig:figure5}
    \vspace{-2mm}
\end{figure*}
}

\newcommand{\pareto}{
\begin{figure}[t]
    \centering
    \includegraphics[width=\linewidth]{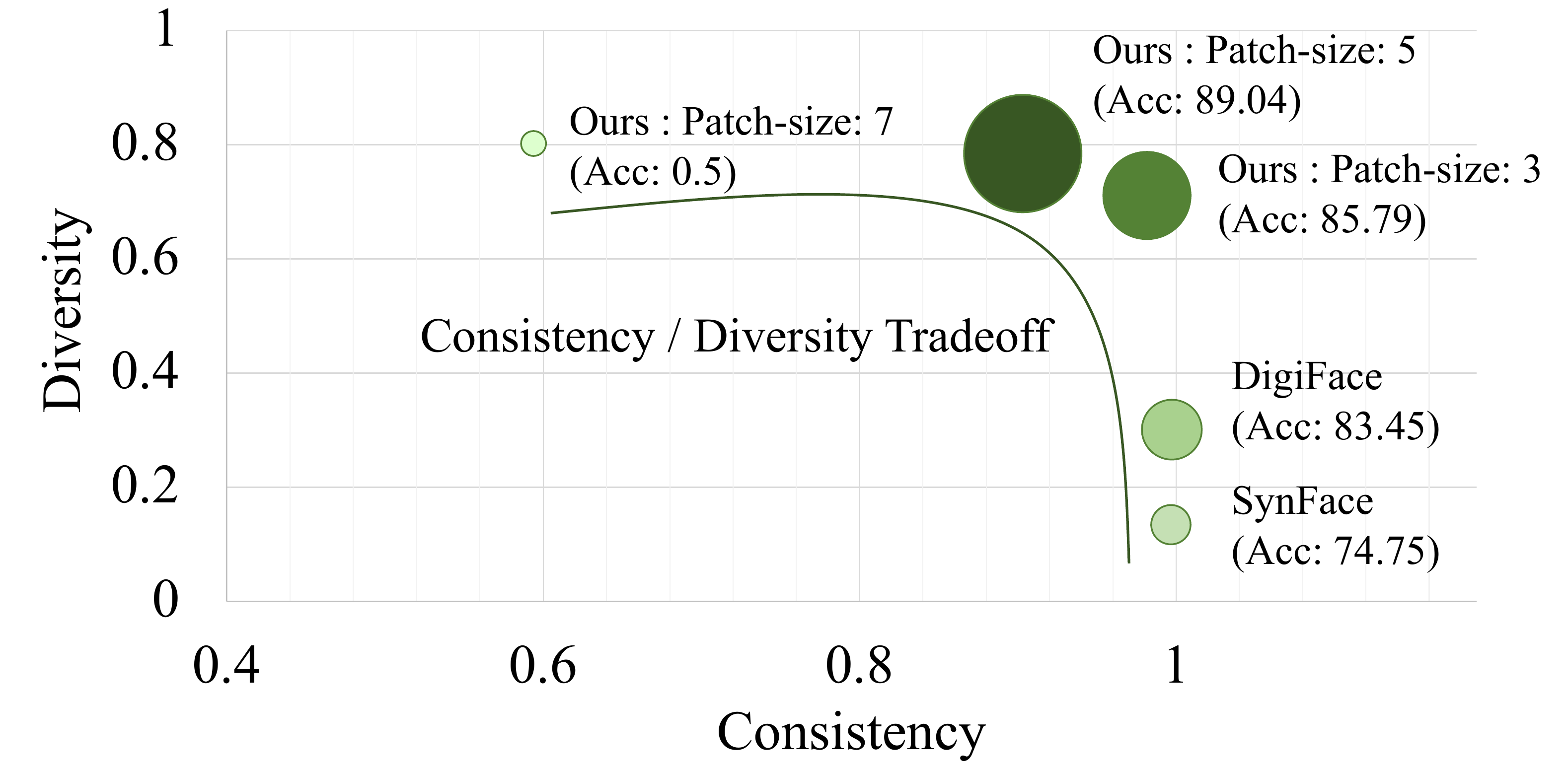}
    \vspace{-6mm}
    \caption{A plot of FR performance on $5$ synthetic datasets with respect to Consistency and Diversity metrics. Color intensity and circle size denotes the FR accuracy. }
    \vspace{-2mm}
    \label{fig:pareto}
\end{figure}
}

\newcommand{\figVisCompare}{
\begin{figure*}[t]
    \centering
    \includegraphics[width=0.95\linewidth]{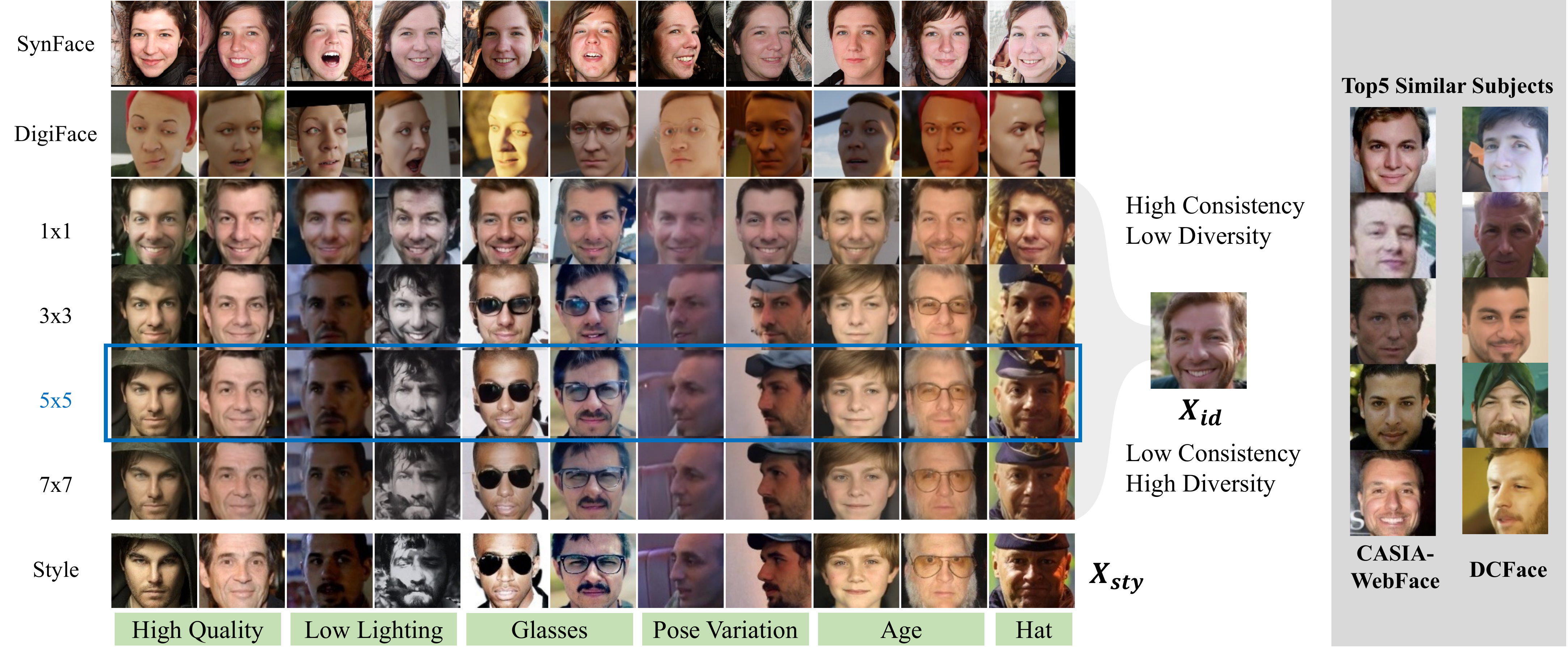}
    \caption{An example of SynFace and DigiFace in rows $1$-$2$ and DCFace with different grid size settings in rows $3$-$7$. SynFace (DiscoFaceGAN) generates mostly frontal-view high-quality images and DigiFace contains synthetic face images with unrealistic texture compared to real images. Our grid size ablation changes the contribution of $\bm{X}_{sty}$ and $\bm{X}_{id}$. A good FR performance is a compromise in-between, $5\!\times\!5$. Note that our method can have diverse styles such as low lighting,  pose, glassses, hat, \textit{etc}. 
    Using $\bm{X}_{id}$ to query subjects in CASIA-WebFace and DCFace datasets returns top $5$ most similar subjects. 
    We see $\bm{X}_{id}$ sufficiently different from other (real or fake) subjects. }
    \label{fig:imagecompare}
\end{figure*}
}

\title{DCFace: Synthetic Face Generation with Dual Condition Diffusion Model}

\author{Minchul Kim\\
{\tt\small kimminc2@msu.edu}
\and
Feng Liu \\
  {\tt\small liufeng6@msu.edu} \\
\and
Anil Jain \\
  {\tt\small jain@msu.edu} \\
\and
Xiaoming Liu \\
  {\tt\small liuxm@msu.edu}\\
\and
  Michigan State University \\
  East Lansing, MI 48824 \\
}
\maketitle

\begin{abstract}
Generating synthetic datasets for training face recognition models is challenging because dataset generation entails more than creating high fidelity images. It involves generating multiple images of same subjects under different factors (\textit{e.g.}, variations in pose, illumination, expression, aging and occlusion) which follows the real image conditional distribution. Previous works have studied the generation of synthetic datasets using GAN or 3D models. In this work, we approach the problem from the aspect of combining subject appearance (ID) and external factor (style) conditions. These two conditions provide a direct way to control the inter-class and intra-class variations. To this end, we propose a Dual Condition Face Generator (DCFace) based on a diffusion model. Our novel Patch-wise style extractor and Time-step dependent ID loss enables DCFace to consistently produce face images of the same subject under different styles with precise control. Face recognition models trained on synthetic images from the proposed DCFace provide higher verification accuracies compared to previous works by $6.11\%$ on average in $4$ out of $5$ test datasets, LFW, CFP-FP, CPLFW, AgeDB and CALFW. \href{https://github.com/mk-minchul/dcface}{Code Link}
\end{abstract}

\vspace{-4mm}
\Section{Introduction}

What does it take to create a good training dataset for visual recognition? An ideal training dataset for recognition tasks would have 1) large inter-class variation, 2) large intra-class variation and 3) small label noise. 
In the context of face recognition (FR), it means, the dataset has a large number of unique subjects, large intra-subject variations, and reliable subject labels. 
For instance, large-scale face datasets such as WebFace4M~\cite{zhu2021webface260m} contain over $1$M subjects and large number of images/subject. 
Both the number of subjects and the number of images per subject are important for training FR models~\cite{deng2019arcface,kim2022adaface}. 
Also, datasets amassed by crawling the web are not free from label noise~\cite{zhu2021webface260m, cao2018vggface2}. 

\figone

In various domains, synthetic  datasets are traditionally used to help generalize deep models when only limited real datasets could be collected~\cite{engelsma2022printsgan,tremblay2018training,zunair2021synthesis,hu2021sail} or when bias exists in the real dataset~\cite{kupas2021solving,van2021decaf}. 
Lately, more attention has been drawn to training with only synthetic datasets in the face domain, as synthetic data can avoid leaking the privacy of real individuals. 
This is important as real face datasets have been under scrutiny for their lack of informed consent, as web-crawling is the primary means of large-scale data collection~\cite{msceleb,zhu2021webface260m,casia}. 
Also, synthetic training datasets can remedy some long-standing issues in real datasets, {\it e.g.}~the long tail distribution, demographic bias, {\it etc.}

\figtwo
When it comes to generating synthetic training datasets, the following questions should be raised. (i) How many novel subjects can be synthesized (ii) How well can we mimic the distribution of real images in the target domain and (iii) How well can we consistently generate multiple images of the same subjects? We start with the hypothesis that face dataset generation can be formulated as a problem that maximizes these criteria together.  

Previous efforts in generating synthetic face datasets touch on one of the three aspects but do not consider all of them together~\cite{qiu2021synface,bae2022digiface}. SynFace~\cite{qiu2021synface} generates high-fidelity face images based on DiscoFaceGAN~\cite{deng2020disentangled}, coming close to real images in terms of FID metric~\cite{fid}. However, we were surprised to find that the actual  number of unique subjects 
that can be generated by DiscoFaceGAN 
is less than $500$, a finding that will be discussed in Sec.~\ref{sec:prelim}. The recent state of the art (SoTA), DigiFace~\cite{bae2022digiface}, can generate $1$M large-scale synthetic face images with many unique subjects based on $3$D parametric model rendering. 
However, it falls short in matching the quality and style of real face images.

We propose a new data generation scheme that addresses all three criteria, {\it i.e.}~the large number of novel subjects (\textit{uniqueness}), real dataset style matching (\textit{diversity}) and label consistency (\textit{consistency}).  In Fig.~\ref{fig:figure1}, we illustrate the high-level idea by showcasing some of our generated face samples. The key motivation of our paper is that the synthetic dataset generator needs to control the number of unique subjects, match the training dataset's style distribution and be consistent in the subject label.

In light of this, we formulate the face image generation as a dual condition inverse problem, retrieving the unknown image $\mathbf{Y}$ from the observable Identity condition $\mathbf{X}_{id}$ and Style condition $\mathbf{X}_{sty}$. Specifically, $\mathbf{X}_{id}$ specifies how a person looks and $\mathbf{X}_{sty}$ specifies how $\mathbf{X}_{id}$ should be portrayed in an image. $\mathbf{X}_{sty}$ contains identity-independent information such as pose, expression, and image quality. 

Our choice of dual conditions (identity and style) is important in how we generate a synthetic dataset as ID and style conditions are controllable factors that govern the dataset's characteristics. To achieve this, we propose a two-stage generation paradigm. First, we generate a high-quality face image $\mathbf{X}_{id}$ using a face image generator and sample a style image $\mathbf{X}_{sty}$ from a style bank. Secondly, we mix these two conditions using a dual condition generator which predicts an image that has the ID of $\mathbf{X}_{id}$ and a style of $\mathbf{X}_{sty}$. An illustration is given in Fig.~\ref{fig:figure2}. 

Training the mixing generator in stage 2 is not trivial as it would require a triplet of ($\mathbf{X}^A_{id}, \mathbf{X}^B_{sty}$, $\mathbf{X}^A_{sty}$) where $\mathbf{X}^A_{sty}$ is a hypothetical combination of the ID of subject $A$ and the style of subject $B$. To solve this problem, we propose a new dual condition generator that can learn from ($\mathbf{X}^A_{id}, \mathbf{X}^A_{sty}$), a tuple of same subject images that can always be obtained in a labeled dataset. The novelty lies in our style condition extractor and ID loss which prevents the training from falling into a degenerate solution. We modify the diffusion model~\cite{sohl2015deep,ho2020denoising} to take in dual conditions and apply  an auxiliary time-dependent ID loss that can control the balance between sample diversity and label consistency. 

We show that our Dual Condition Face Dataset Generator (DCFace) is capable of surpassing the previous methods in terms of FR performance, establishing a new benchmark in face recognition with synthetic face datasets. We also show the roles dataset subject uniqueness, diversity and consistency play in face recognition performance. 

The followings are the contributions of the paper.
\begin{itemize}
    \item We propose a two-stage face dataset generator that controls subject uniqueness, diversity and consistency. 
    \item For this, we propose a dual condition generator that mixes the two independent conditions $\mathbf{X}_{id}$ and $\mathbf{X}_{sty}$. 
    \item We propose uniqueness, consistency and diversity metrics that quantify the respective properties of a given dataset, useful measures that allow one to compare datasets apart from the recognition performance.
    \item We achieve SoTA in FR with $0.5$M image synthetic training dataset by surpassing the previous methods by $6.11\%$ on average in $5$ popular test datasets.
    
\end{itemize}
\Section{Related Works}

\Paragraph{Face Recognition}
Face Recognition (FR) is the task of matching query imagery to an enrolled identity database. SoTA FR models are trained on large-scale web-crawled datasets~\cite{zhu2021webface260m,msceleb,deng2019arcface} with margin-based softmax losses~\cite{wang2018cosface, deng2019arcface, liu2017sphereface, huang2020curricularface,kim2022adaface}. The FR performance is measured on various benchmark datasets such as LFW~\cite{lfw}, CFP-FP~\cite{cfpfp}, CPLFW~\cite{cplfw}, AgeDB~\cite{agedb} and CALFW\cite{calfw}. 
These datasets are designed to measure factors such as  pose changes and age variations.
Performance on these datasets for models trained on large-scale datasets such as WebFace260M is well above $97\%$~\cite{kim2022adaface} in  verification accuracy. 

\Paragraph{Synthetic Face Generation} 
Recent advances in generative models allow high fidelity synthetic face image generations~\cite{karras2019style,choi2018stargan,karras2017progressive,karras2020analyzing,brock2018large,ho2020denoising,song2020denoising}.
GANs have been widely used to manipulate, animate or enhance face images
~\cite{hu2018disentangling,deng2020disentangled,xiao2018elegant,pumarola2018ganimation,sun2019single,choi2018stargan,lin2018conditional,disentangled-representation-learning-gan-for-pose-invariant-face-recognition}. They typically learn  disentangled representations in GAN latent space that control desired face properties. On the contrary, some works leverage the 3D face prior from 3D datasets (\emph{e.g.}, $3$DMM~\cite{blanz1999morphable}) for controllable synthesis~\cite{shen2018facefeat,kim2018deep,deng2018uv,gecer2018semi,geng20193d,piao2019semi,nguyen2019hologan,most-gan-3d-morphable-stylegan-for-disentangled-face-image-manipulation}. These methods have advantages in the fine-grained control over face generation and 3D consistency yet lack in style or domain variation. 

Recent advances in the latent variable models such as diffusion or score-based models have shown great success in high-quality image generation with a more stable and simple objective of MSE loss~\cite{ho2020denoising,nichol2021improved,sohl2015deep,song2019generative,song2020improved,song2021maximum,song2020denoising}. Diffusion models have advanced the conditional image generation in tasks such as text-conditional image generation, inpainting, {\it etc}~\cite{piti,dalle2,blattmann2022retrieval,rombach2022high}. We adopt the diffusion model as a backbone and explore how the two image characteristics, namely ID and style images, can control complementary information, the subject appearance and the style of an image.

\Paragraph{Face Recognition with Synthetic Dataset}
Synthetic training datasets offer an advantage over real datasets with regards to ethical issues and class imbalance problems as large-scale face datasets have been criticized for lacking informed consent and reflecting racial biases~\cite{zhu2021webface260m,deng2019arcface,yi2014learning,bae2022digiface}. Despite the benefit, use of synthetic datasets as the sole training data is not widely adopted due to the resulting low recognition performance. In various domains such as face recognition~\cite{qiu2021synface,bae2022digiface,controllable-and-guided-face-synthesis-for-unconstrained-face-recognition},  fingerprint recognition~\cite{engelsma2022printsgan,wyzykowski2022synthetic}, and anti-spoofing~\cite{9779478,noise-modeling-synthesis-and-classification-for-generic-object-anti-spoofing}, synthetic datasets have been shown to improve recognition when combined with real images. 

In the face domain, SynFace~\cite{qiu2021synface} studied the efficacy of using DiscoFaceGAN~\cite{deng2020disentangled} for synthetic face generation. Recently, DigiFace-1M~\cite{bae2022digiface} studied the efficacy of 3D model based face rendering in combination with image augmentations to create a synthetic dataset. We propose a face dataset generation method that can generate both a large number of subjects and diverse styles that are close to the real dataset.  

\Section{Proposed Approach}
We propose Dual Condition Face Dataset Generator (DCFace), a two-stage dataset generator (see Fig.~\ref{fig:figure2}). 
Stage $1$ is the Condition Sampling Stage, generating a high-quality ID image ($\mathbf{X}_{id}$) of a novel subject and selects one arbitrary style image ($\mathbf{X}_{sty}$) from the bank of real training data. Stage $2$ is the Mixing Stage which combines the two images using the Dual Condition Generator. 

For trainable models in each stage, Stage $1$ requires training an ID image generator $G_{id}$.  
For the style bank, we can conveniently use any real face dataset that we wish generated samples to follow. Stage $2$ requires training a dual condition mixer $G_{mix}$. Both $G_{id}$ and $G_{mix}$ are based on diffusion models~\cite{ho2020denoising}. We describe each component and the associated training procedure in the following subsections. 


\figGANDDPM 

\algorithm

\subsection{Preliminary}
\label{sec:prelim}
Diffusion models~\cite{sohl2015deep,ho2020denoising} are a class of denoising generative models that are trained to predict an image from random noise through a gradual denoising process. One notable difference from the class of GAN-based generators~\cite{goodfellow2020generative} is in the objective function and the sampling procedure. The forward process as expressed in Eq.~\ref{eq:diff} corrupts the input $\bm{X}$ using variance controlled Gaussian noise over $t$ time-steps,
\vspace{-1mm}
\begin{equation}
    q\left(\bm{\bm{X}}_{t}|\bm{\bm{X}}_{t-1}\right)=\mathcal{N}\left(\bm{\bm{X}}_{t};\sqrt{1-%
\beta_{t}}\bm{\bm{X}}_{t-1},\beta_{t}\bm{I}\right), 
\label{eq:diff}
\end{equation}
\vspace{-2mm}

\noindent and the denoising is done by training a model $\bm{\epsilon}_{\theta}(\bm{X}_{t},t)$ to predict the initial noise $\bm{\epsilon}$ with an $L_2$ objective, 
\vspace{-2mm}
\begin{equation}
\mathcal{L}=\mathbb{E}_{t,\bm{X}_{0},\bm{\epsilon}}\Big{[}\big{\|}\bm{\epsilon}_{\theta}(\underbrace{\sqrt{%
\alpha_{t}}\bm{X}_{0}+\sqrt{1-\alpha_{t}}\bm{\epsilon}}_{\bm{X}_{t}},t)-\bm{\epsilon}\big{\|}_{2}^{2}\Big{]}.
\label{eq:ddpm}
\end{equation}
\vspace{-2mm}

\noindent $\beta_{t}$ and $\alpha_{t}$ are pre-set variance scheduling scalars.
The denoising diffusion model (DDPM) has shown success in producing diverse samples in text-conditioned image generation~\cite{dalle2}. We find that in unconditional face generation, DDPM is also capable of generating many unique subjects. For instance, Fig.~\ref{fig:ganddpmcomp} compares DiscoFaceGAN~\cite{deng2020disentangled} with DDPM~\cite{ho2020denoising} in their capacity to generate different subjects for every sample. It shows that DDPM~\cite{ho2020denoising} is a good model choice for $G_{id}$ and $G_{mix}$ as it can generate many unique subjects. For $G_{id}$, we adopt the unconditional DDPM trained on FFHQ~\cite{karras2019style}, having observed that it is capable of generating a large number of unique subject images.    

\subsection{Dual Condition Generator $G_{mix}$ }
The two-stage data generation requires Dual Condition Generator $G_{mix}$ which is a conditional DDPM. 
Two conditions $\bm{X}_{id}$ and $\bm{X}_{sty}$ are injected into the denoiser $\bm{\epsilon}_{\theta}(\bm{X}_{t},t,E_{id}(\bm{X}_{id}), E_{sty}(\bm{X}_{sty}))$ using trainable feature extractors $E_{id}$ and $E_{sty}$ and cross-attentions. $G_{mix}$ is responsible for the operation $\bm{X}_{id}^A+\bm{X}_{sty}^B\!\rightarrow\!\bm{X}_{sty}^A$, a mixing of an image of a novel subject $A$ and an arbitrary style image of different subject $B$. 

Naive training would require the reference image $\bm{X}_{sty}^A$, an image of subject $A$ in the style of $\bm{X}_{sty}^B$. This reference is absent in the labeled training dataset. As such, we modify the operation to $\bm{X}_{id}^A+\bm{X}_{sty}^A\!\rightarrow\!\bm{X}_{sty}^A$, using two different images from the same subject as illustrated in Fig.~\ref{fig:algorithm}(a). But this formulation is prone to a trivial solution of ignoring $\bm{X}_{id}^A$, making the ID condition unused during test time. To mitigate this issue, we propose the following two elements.   

\Paragraph{Patch-wise Style Extractor $E_{sty}$} The motivation of Style Extractor is to map an image $\bm{X}_{sty}$ to a feature that contains little ID information, forcing $G_{mix}$ to rely on $\bm{X}_{id}$ for ID information. In prior works such as StyleGAN, $1^{st}$ and $2^{nd}$ order statistics of a feature are shown to resemble the image style~\cite{karras2019style,caface,lee2019srm}. Yet, resulting statistics are reduced in spatial dimensions and consequently without spatially local informations such as pose.   

We propose a module that can extract style information without losing spatial information. Specifically, consider a pretrained and fixed face recognition model $F_{s}$ and its intermediate feature $F_{s}(\mathbf{X}_{sty})=\mathbf{I}_{sty} \!\in\! \mathbb{R}^{C\!\times\!H\!\times\!W} $. We divide the feature into a $k\times\!k$ grid. For each element in the grid $\mathbf{I}_{sty}^{k_i}\!\in\!  \mathbb{R}^{C\!\times\!\frac{H}{k}\!\times\!\frac{W}{k}}$, we perform non-linear mapping on the mean and variance of $\mathbf{I}_{sty}^{k_i}$. Specifically,   
\begin{gather}
\mathbf{\hat{I}}^{k_i} = \text{BN}(\text{Conv}(\text{ReLU}(\text{Dropout}(\mathbf{I}_{sty}^{k_i})))),\\
\bm{\mu}_{\text{sty}}^{k_i} = \text{SpatialMean}(\mathbf{\hat{I}}^{k_i}), \quad  \bm{\sigma}_{\text{sty}}^{k_i} = \text{SpatialStd}(\mathbf{\hat{I}}^{k_i}), \\
\bm{s}^{k_i} = \text{LN}\left((\bm{W}_1 \odot \bm{\mu}_{\text{sty}}^{k_i} + \bm{W}_2 \odot\bm{\sigma}_{\text{sty}}^{k_i}) + \bm{P}_{emb} \right),\\
E_{sty}(\bm{X}_{sty}) := \bm{s} = [\bm{s}^1, \bm{s}^2, \bm{s}^{k_i}..., \bm{s}^{k\!\times\!k}, \bm{s}'],
\end{gather}
where $\bm{s}'$ corresponds to $\mathbf{I}_{sty}^{k_i}$ being a global feature, where $k\!=\!1$.
The final output $\bm{s}$ is a concatenation of all style vectors for each patch.  Each $\bm{s}^{k_i}$ is a mean and variance of local information which is constrained from containing full pixel-level details with the ID information. And $\bm{P}_{emb}$ is a learned position embedding to let the model differentiate different patch locations. BN and LN are BatchNorm~\cite{ioffe2015batch} and LayerNorm~\cite{ba2016layer}. $F_{s}$ is a shallow CNN taken from the early layers of a pretrained FR model. It is fixed and not updated to prevent it from optimizing $\mathbf{I}_{sty}$, serving only to create style information. By varying the grid size $k\!\times\!k$, we can represent style at different spatial locations. An illustration of $E_{sty}$ can be found in Fig.~\ref{fig:algorithm}(b).


\figfive

\Paragraph{Time-step Dependent ID Loss}
To train Dual Condition Generator $G_{mix}$, the original DDPM objective of $L_2$ loss, Eq.~\ref{eq:ddpm} is not sufficient to guarantee the consistency in subject identity between the ID condition $\bm{X}_{id}$ and the prediction, $\hat{\bm{X}}_{0}$. To ensure the ID consistency, one could devise a loss function to maximize the similarity between $\bm{X}_{id}$ and the predicted denoised image $\hat{\bm{X}}_{0}$, in the ID feature space using a pretrained FR model, $F$. Specifically, following the Eq.15 of DDPM~\cite{ho2020denoising}, one-step prediction of the original image is 
\vspace{-4mm}
\begin{gather}
\hat{\bm{X}}_{0} = (\bm{X}_t - \sqrt{1-\bar{\alpha}_t} \bm{\epsilon}_{\theta}(\bm{X}_{t},t,\bm{X}_{id}, \bm{X}_{sty}) ) / \sqrt{\bar{\alpha}_t}.
\end{gather}
A simple ID loss to increase cosine similarity (CS) is
\begin{equation}
        L_{\text{naive1}} = -\text{CS}\left(F(\bm{X}_{id}) , F(\hat{\bm{X}}_{0}) )\right).
\end{equation}
However, this loss is in conflict with MSE loss and is empirically observed to reduce the predicted image quality. This is because the FR model, $F$ is not invariant to image style; some style of $\bm{X}_{id}$ has to match in order to completely reduce $L_{\text{naive1}}$. 
In contrast, one could also use 
\begin{equation}
        L_{\text{naive2}} = -\text{CS}\left(F(\bm{X}_{sty}) , F(\hat{\bm{X}}_{0}) )\right),
\end{equation}
as during training the label of $\bm{X}_{sty}$ and $\bm{X}_{id}$ are the same. However, $L_{\text{naive2}}$ causes the model to depend on $\bm{X}_{sty}$ for ID information. Thus, during evaluation, when $\bm{X}_{sty}$ and $\bm{X}_{id}$ are different subjects, the label consistency in the generated dataset is compromised. We show this in Tab.~\ref{table:modelAblation}.

Instead, we propose to interpolate between $F(\bm{X}_{id})$ and $F(\bm{X}_{sty})$ across diffusion time-steps. Specifically,
\vspace{-2mm}
\begin{equation}
\begin{split}
\label{eq:lid}
        L_{\text{ID}} = &-\gamma_t \text{CS}\left(F(\bm{X}_{id}) , F(\hat{\bm{X}}_{0}) )\right) \\
        &- (1-\gamma_t) \text{CS}\left(F(\bm{X}_{sty}) , F(\hat{\bm{X}}_{0}) )\right),
        \vspace{-3mm}
\end{split}
\end{equation}
\vspace{-2mm}

\noindent where $\gamma_t\!=\!\frac{t}{T}$ is a time-dependent weight that linearly changes from $0$ to $1$. When $t\!=\!T$, $\bm{\epsilon}_\theta$ is predicting $\bm{X}_{t-1}$ from random noise, and we let the model fully exploit the ID information of $\bm{X}_{id}$.  Gradually as $t$ increases, we let the model's prediction walk into the direction of $\bm{X}_{sty}$. Note that during training, the actual label of $\bm{X}_{sty}$ and $\bm{X}_{id}$ are the same. So the interpolation in the loss forces the prediction to be the same in identity but gradually shifting in style toward $\bm{X}_{sty}$. This loss allows $\bm{\epsilon}_{\theta}(\bm{X}_{t},t,\bm{X}_{id}, \bm{X}_{sty}))$ to play different roles depending on $t$. For $t\approx T $, $\bm{\epsilon}_{\theta}$ will exploit $\bm{X}_{id}$ to infer front-view ID rich image. And as $t\rightarrow 0$, it will change the image's style to match the style of $\bm{X}_{sty}$.    
The final loss is $L_{MSE}+\lambda L_{ID}$ with $\lambda$ as a scaling parameter.

\Paragraph{$\bm{E}_{id}$ and Conditioning Mechanism} 
Following the success text-conditional image generation and inpainting using DDPM~\cite{dalle2,piti,diffae}, we adopt a similar architecture for inserting conditions into the model. We concatenate $E_{id}(\bm{X}_{id})$ and $E_{sty}(\bm{X}_{sty})$ and put in $\bm{\epsilon}_{\theta}$ using cross-attention and  adaptive
group normalization layers (AdaGN)~\cite{diffae}. $E_{id}$ is a CNN, with the same architecture as a small FR model (\textit{e.g.} ResNet50). And $E_{id}$ is trained end-to-end with $\bm{\epsilon}_\theta$ to extract useful ID feature for $\bm{\epsilon}_\theta$. 
More training details can be found in Supp.

\subsection{Condition Sampling Strategy}
\label{sec:sampling}
\Paragraph{ID Image Sampling} For sampling ID images, we generate $200,000$ facial images from $G_{ID}$, from which we remove faces that are wearing sunglasses or too similar to the subjects in CASIA-WebFace with the Cosine Similarity threshold of $0.3$ using $F_{eval}$. We are left with $105,446$ images. Then we narrow them down to $62,570$ images that are unique according to uniqueness, Eq.~\ref{eq:unique} using $F_{eval}$ and $r=0.3$. Then we explore two different options, 1) random sampling and 2) gender/ethnicity balanced sampling as $G_{id}$ has a skewed distribution towards White subjects as shown in Tab.~\ref{tab:distribution}. 
We use~\cite{faceparsing} to classify the ethnicity and use~\cite{glasses,glassesgit} to detect sunglasses. 
We denote the sampling option 1 as $\textit{random}$ and 2 as $\textit{balance}$.

\Paragraph{Style Image Sampling}
For style sampling, for each $\mathbf{X}_{id}$, we randomly sample $\mathbf{X}_{sty}$ from the style bank. We denote this option as $\textit{random}$. We also explore the option of sampling $\mathbf{X}_{sty}$ from the pool of images whose gender/ethnicity matches that of $\mathbf{X}_{id}$. We denote this option as $\textit{match}$.

\ethnicityratio

\Section{Dataset Evaluation}

In evaluating the synthesized dataset, one often adopts 1) FID~\cite{fid} for evaluating the distribution similarity to the real images and 2) subsequent recognition performance. In this section, we propose three class-dependent metrics that aid us in understanding the property of generated labeled datasets. We let $F_{eval}$ be an recognition model used for evaluating synthesized face datasets. Note that this is different from $F$ in ID loss. $F$ is a model for training loss and $F_{eval}$ is for evaluating metrics. The more generalizable $F_{eval}$ is, the more accurate the metrics become in capturing the identity and diversity of the synthesized dataset.

Let $y_c$ be a class label, and $f_i=F_{eval}(\bm{X}_i)$. Let $d(f_i, f_j)$ be the distance between two images in $F_{eval}$ feature space. 

\Paragraph{Uniqueness} 
Consider the following non-overlapping $r$-ball in $F_{eval}$ space, 
\begin{equation}
    U\!=\!\{ f_i : d(f_i, f_j)>r,\; j < i, \; i,j\in \{1,..,N\}\},
    \label{eq:unique}
\end{equation}
where  $d(f_i, f_j)$ is the cosine distance.
Then $|U|$ is the count of unique subjects determined by the threshold $r$ in an unlabeled dataset. Note that the set $U$ is equivalent to sequentially adding a $r$-ball into $F_{eval}$-space until you cannot add more without collision. $|U|$ is subject to both $r$ and $F_{eval}$. In FR, $r$ is a threshold in the FR model that is set to determine match or non-match. 

For a labeled synthetic dataset, one generates multiple feature sets $\{f_i^c\}$ for the same label. To count the number of unique subjects, we calculate the number of unique centers, $ f^c = \frac{1}{N_c} \sum_{i}^{N_c} f_i^c$ for $c\in\{1,...,C\}$, where $C$ is the number of subjects and $N_c$ is the number of images per subject. Then we define the number of unique subjects in a labeled dataset with $|U_c|$ where $U_c$ is
\begin{equation}
    U_c\!=\!\{ f_c \!:\! d(f^{c_n}, f^{c_m})\!>\!r, m\! <\! n, n,m\!\in\! \{1,..,C\}\},
    \label{eq:uniqueclass}
\end{equation}
For the metric, we use $U_{class} = |U_c| / C$, the ratio between the number of unique subjects and the number of labels.  

\Paragraph{Intra-class Consistency} It measures how consistent the generated samples are in adhering to the label condition, as 
\vspace{-2mm}
 \begin{equation}
  C_{intra} = \frac{1}{C} \sum_{c=1}^C \frac{1}{N_c} \sum_{i=1}^{N_c} d(f_i^c, f^c) < r, 
 \end{equation}
 \vspace{-2mm}
 
\noindent which is the ratio of individual features $f_i^c$ being close to the class center $f^c$. For a given threshold $r$, higher values of $C_{intra}$ mean the samples are more likely to be the same subject under the same label.


\Paragraph{Intra-class Diversity}
It measures how diverse the generated samples are under the same label condition. Note that the diversity is in the style of an image, not in the subject's identity. We define the style space as a vector space defined by Inception Network~\cite{salimans2016improved} features pretrained on ImageNet~\cite{deng2009imagenet} following the convention of~\cite{kynkaanniemi2019improved}, denoting the real and generated image inception vectors as $\{s_i^c\}$, $\{\hat{s}_j^c\}$. 

For intra-class diversity, we measure how many real images fall into the style space manifold defined by the generated images under the same label condition. 
We compute this by extending the Improved Recall Metric~\cite{kynkaanniemi2019improved}, from comparing the unconditional distributions of real and fake images to comparing the label-conditional distributions.
Specifically, for a set of real and generated feature vectors $\{s_i^c\}$, $\{\hat{s}_j^c\}$ under the same label condition $y_c$, we define $k$-nearest feature distance $r_k$ as  
    $r_k = d\bigl(%
\hat{s}_j^c -\text{NN}_{k}\left(\hat{s}_j^c,
\{\hat{s}_j^c\} \right)\bigr)$,
where $\text{NN}_{k}$ returns the $k$-nearest feature vector in $\{\hat{s}_j^c\}$ and
\vspace{-2mm}
\begin{equation}
\small
    \mathbf{I}(s_i^c, \{\hat{s}_j^c\}
 )\!=\!\begin{cases}1, \exists \hat{s}_j^c \in \{\hat{s}_j^c\} \text{ s.t. }d\left(s_i^c-\hat{s}_j^c\right)\leq r_k  \\
0,\text{ otherwise}. \end{cases}
\label{eq:indicator}
\end{equation}
\vspace{-2mm}

\noindent $d(\cdot)$ is an Euclidean distance. Then diversity is defined by 
\vspace{-2mm}
\begin{equation}
\text{D}_{intra} = \frac{1}{C} \frac{1}{N} \sum_{c=1}^C \sum_{i=1}^{N_c} \mathbf{I}(s_i^c, \{\hat{s}_j^c\}),
\end{equation} 
\vspace{-2mm}

\noindent which is the fraction of real image styles manifold covered by the generated image style manifold as defined by $k$-nearest neighbor ball. If the style variation is small, then $r_k$ becomes small, reducing the chance of $d\left(s_i^c-\hat{s}_j^c\right)\leq r_k$. We compute the recall per class to capture style variation conditional on the subject label. 

In Fig.~\ref{fig:figure5}, we illustrate different scenarios of conditional generation and how these metrics can capture the shortcomings in each scenario. 
In Sec.~\ref{sec:exp} and Fig.~\ref{fig:pareto}, we measure the metrics on our generated datasets and compare with previous synthetic datasets~\cite{qiu2021synface,bae2022digiface}. 
We find that FR performance is at best when consistency and diversity are balanced. 
Also, we find SynFace and DigiFace have high $C_{intra}$ and low $D_{intra}$ compared to our method in Fig.~\ref{fig:figure5}. 

\pareto

\Section{Experiments}
\label{sec:exp}

For $G_{id}$ which generates ID images, we adopt the publicly released unconditional DDPM~\cite{ho2020denoising} trained on FFHQ~\cite{karras2019style}. For $G_{mix}$, we train it on CASIA-WebFace~\cite{casia} after initializing weights from $G_{id}$. Although using all of CASIA-WebFace is a valid setting, we split it into a $95$-$5$ split between train and validation sets. 
The validation set is used as a real dataset in measuring the uniqueness, consistency and diversity metrics.
$G_{mix}$ is trained for $10$ epochs with a batch-size of $256$ using AdamW Optimizer~\cite{kingma2014adam,loshchilov2017decoupled} with the learning rate of $0.001$. Training takes $8$ hours using two A100 GPUs.
Once $G_{mix}$ is trained, we use $G_{id}$, $G_{mix}$ and a style bank to generate a synthetic labeled dataset. 
The style bank is the CASIA-WebFace training set. For sampling, we use DDIM~\cite{song2020denoising} with $200$ intervals. Generating $500$K samples takes about $20$ hours using one A100 GPU. 

To train FR models, for a fair comparison, we adopt the training scheme of~\cite{bae2022digiface,qiu2021synface} using IR-SE-50~\cite{deng2019arcface} as a backbone and AdaFace~\cite{kim2022adaface} as a loss function. We evaluate the trained FR models on five datasets, LFW~\cite{lfw}, CFP-FP~\cite{cfpfp}, CPLFW~\cite{cplfw}, AgeDB~\cite{agedb} and CALFW\cite{calfw}.  
CFP-FP and CPLFW are designed to measure the FR in the large pose variation and AgeDB and CALFW are for the large age variation. 
To measure the consistency, diversity and uniqueness during evaluation, we adopt $F_{eval}$ as IR101~\cite{deng2019arcface} model trained on WebFace4M~\cite{zhu2021webface260m} with AdaFace~\cite{kim2022adaface} loss. 


\modelAblation

\subsection{Model Ablation}
\label{sec:ablation}
To show the efficacy of our proposed modules, we ablate on 1) the grid size in Style extractor $E_{sty}$, 2) Time-step dependent ID loss and 3) the ID loss backbone $F$'s.
The number of samples we generate for the ablation are $10K$ subjects with $50$ images per subject, similar to CASIA-WebFace image counts. 
We report the FR performance with the synthetic data by averaging the $5$ validation set verification accuracies. 
To measure $U_{class}$, $C_{intra}$ and $D_{intra}$, we use $500$ subjects with $20$ real images from the held-out validation set of CASIA-WebFace and generate an equivalent number of images from each method. 

\Paragraph{Grid Size} We choose $4$  grid sizes ranging from $1\!\times\!1$ to $7\!\times\!7$. Note that  $1\!\times\!1$ corresponds to the style vector of a whole image. 
We expect to see higher spatial control in $\bm{X}_{sty}$ as the grid size increases. 
In Tab.~\ref{table:modelAblation}, we report the three metrics $U_{class}$, $C_{intra}$ and $D_{intra}$. 
As the grid size increases, $E_{sty}$ features contain more fine-grained information, possibly related to ID, lowering the consistency. 
However, the diversity increases, making the conditional distribution similar to the real dataset. 
The subsequent FR performance using the model is the best in the setting $5\!\times\!5$, which is a good compromise between consistency and diversity. 
In Fig.~\ref{fig:imagecompare}, we show the effect of the grid size with examples. 

\Paragraph{ID Loss}
For ID loss, we compare $L_{\text{ID}}$ with $L_{\text{naive1}}$ and $L_{\text{naive2}}$ in Tab.~\ref{table:modelAblation}. Using $L_{\text{naive1}}$ or $L_{\text{naive2}}$ both suffer from lower FR performance, but for different reasons. $L_{\text{naive1}}$ has low diversity because it is optimized to be similar to $\bm{X}_{id}$ of front-view high quality face images. 
$L_{\text{naive2}}$ has low consistency because of the lack of dependence on $\bm{X_}{id}$, making the resulting dataset with random labels. FR performance of $0.5$ means the model diverged and is returning random predictions. $L_{\text{ID}}$, a linear interpolation of the $L_{\text{naive1}}$ and $L_{\text{naive2}}$ across time-steps results in the best performance. 

\Paragraph{ID Loss Backbone $\bm{F}$} ID Loss requires a pretrained FR model, $F$. 
For all of our experiments, we use $F$ as IR50 trained on CASIA-WebFace. 
But, we are curious if there is a benefit to have a better representation from $F$. 
For this, we ablate $F_{bigger}$, a model pretrained on a larger dataset, WebFace4M~\cite{zhu2021webface260m}. 
Tab.~\ref{table:modelAblation} shows that a better FR backbone induce the generator to synthesize better datasets, even without explicitly showing WebFace4M images to generators. 
But for fairness in comparing to the real CASIA-WebFace dataset, we do not use $\bm{F}_{bigger}$ for subsequent analysis.

\subsection{Sampling Ablation}
Using the sampling strategy defined in Sec.~\ref{sec:sampling}, we ablate on the ID sampling options ($\textit{random}$, $\textit{balance}$) and style sampling methods ($\textit{random}$, $\textit{match}$) in Tab.~\ref{table:samplingAblation}. We find that 
either balancing the gender/ethnicity distribution or making the gender/ethnicity of style image equal to that of ID images does not bring significant performance gain.

\samplingAblation

\figVisCompare

\benchmarkTable

On the other hand, to compensate for lower label consistency compared to the real dataset, we include the same $\bm{X}_{id}$ for $5$ additional times for each label. This has the effect of oversampling $\bm{X}_{id}$ during training FR model. When we add the oversampling option to ($\textit{balance}$, $\textit{match}$) setting, we observe an average verification accuracy of $89.56\%$, $0.52\%$ increase over the ($\textit{random}$, $\textit{random}$) setting.

\subsection{Comparison with Previous Methods}
For training FR models with synthetic datasets, we compare with SynFace~\cite{qiu2021synface} and DigiFace~\cite{bae2022digiface}. We compare $0.5$M and $1.2$M image count settings. The first setting corresponds to the size of the CASIA-WebFace real dataset. The second setting is to evaluate the effect of increasing  the training dataset size. In Tab.~\ref{tab:comparison}, we show the verification accuracies of $5$ validation sets. In $0.5$M regime, our DCFace can surpass DigiFace in $4$ out of $5$ datasets with an improvement of $6.11\%$ on average. In CFP-FP dataset with extremely large pose variation, DigiFace performs better, showing the merit of 3D consistent face synthesis using 3D models. DCFace has a good balance of consistency and diversity with many unique subjects, leading to a better FR performance in general.
Note the larger style variation compared to SynFace and DigiFace in Fig.~\ref{fig:imagecompare}. 

The last column of Tab.~\ref{tab:comparison} shows the gap between synthetic and real, calculated as $(\text{REAL}\!-\!\text{SYN})/\text{SYN}$, {\it e.g.} $5.65\%\!=\!\frac{94.62-89.56}{89.56}$. It indicates how much improvement is needed to be on par with the real dataset. In $0.5$M setting, DCFace reduces the gap to real performance by $57\%$ over the SoTA.   
When we use more synthetic data as in $1.2$M regime, the synthetic dataset performance 
comes closer to that of the real dataset ($3.74\%$ in gap), a  $60.9\%$ improvement from the previous method ($9.55\%$ in gap).

\Section{Conclusion}
This paper presents a method for creating a synthetic training dataset for face recognition. 
Dataset generation is studied from the perspective of generating many unique subjects with large style diversity and label consistency. 
We propose the Dual Condition Face Generator to this end and show its large FR performance gain over previous methods on synthetic dataset generation. 
We believe our approach takes one step towards matching the performance of real training datasets with synthetic training datasets. 

\Paragraph{Limitations}
This work addresses the problem of generating label consistent and diverse datasets for face recognition model training. In our model ablation, we find that sacrificing label consistency for diversity to some degree is beneficial for the FR model training. However, this is not ideal; for instance, our synthetic face generator lacks 3D consistency across pose, which is an advantage of generative models with 3D priors. Secondly, the goal of our research is to release a synthetic face dataset that alleviates the dependence on large-scale web-crawled images. As shown in our experiments, there is still some performance gap between real and synthetic training datasets. In this work, we take one step towards the goal and hope that the continued research will introduce a standalone synthetic face dataset. 

\Paragraph{Acknowledgments} This research is based upon work supported in part by the Office of the Director of National Intelligence (ODNI), Intelligence Advanced Research Projects Activity (IARPA), via 2022-21102100004. The views and conclusions contained herein are those of
the authors and should not be interpreted as necessarily representing the official policies, either expressed or implied, of ODNI, IARPA, or the U.S. Government. The U.S. Gov. is authorized to reproduce and distribute reprints for governmental purposes notwithstanding any copyright annotation therein.
\newpage

{\small
\bibliographystyle{ieee_fullname}
\bibliography{egbib}

\begin{thebibliography}{10}\itemsep=-1pt

\bibitem{TFace}
{{TFace}}.
\newblock \url{https://github.com/Tencent/TFace.git}.
\newblock Accessed: 2021-10-3.

\bibitem{faceparsing}
Vítor Albiero.
\newblock Face analysis pytorch.
\newblock \url{https://github.com/vitoralbiero/face_analysis_pytorch}, 2022.

\bibitem{proactive}
Vishal Asnani, Xi Yin, Tal Hassner, Sijia Liu, and Xiaoming Liu.
\newblock Proactive image manipulation detection.
\newblock In {\em CVPR}, 2022.

\bibitem{ba2016layer}
Jimmy~Lei Ba, Jamie~Ryan Kiros, and Geoffrey~E Hinton.
\newblock Layer normalization.
\newblock {\em arXiv preprint arXiv:1607.06450}, 2016.

\bibitem{bae2022digiface}
Gwangbin Bae, Martin de La~Gorce, Tadas Baltrusaitis, Charlie Hewitt, Dong
  Chen, Julien Valentin, Roberto Cipolla, and Jingjing Shen.
\newblock Digiface-1m: 1 million digital face images for face recognition.
\newblock In {\em WACV}, 2023.

\bibitem{blanz1999morphable}
Volker Blanz and Thomas Vetter.
\newblock A morphable model for the synthesis of 3{D} faces.
\newblock In {\em SIGGRAPH}, 1999.

\bibitem{blattmann2022retrieval}
Andreas Blattmann, Robin Rombach, Kaan Oktay, and Bj{\"o}rn Ommer.
\newblock Retrieval-augmented diffusion models.
\newblock {\em arXiv preprint arXiv:2204.11824}, 2022.

\bibitem{brock2018large}
Andrew Brock, Jeff Donahue, and Karen Simonyan.
\newblock Large scale gan training for high fidelity natural image synthesis.
\newblock {\em arXiv preprint arXiv:1809.11096}, 2018.

\bibitem{cao2018vggface2}
Qiong Cao, Li Shen, Weidi Xie, Omkar~M Parkhi, and Andrew Zisserman.
\newblock {VGGFace2}: A dataset for recognising faces across pose and age.
\newblock In {\em FG}, 2018.

\bibitem{tinyface}
Zhiyi Cheng, Xiatian Zhu, and Shaogang Gong.
\newblock Low-resolution face recognition.
\newblock In {\em ACCV}, 2018.

\bibitem{choi2018stargan}
Yunjey Choi, Minje Choi, Munyoung Kim, Jung-Woo Ha, Sunghun Kim, and Jaegul
  Choo.
\newblock Star{GAN}: Unified generative adversarial networks for multi-domain
  image-to-image translation.
\newblock In {\em CVPR}, 2018.

\bibitem{deng2018uv}
Jiankang Deng, Shiyang Cheng, Niannan Xue, Yuxiang Zhou, and Stefanos
  Zafeiriou.
\newblock {UV-GAN}: Adversarial facial uv map completion for pose-invariant
  face recognition.
\newblock In {\em CVPR}, 2018.

\bibitem{deng2009imagenet}
Jia Deng, Wei Dong, Richard Socher, Li-Jia Li, Kai Li, and Li Fei-Fei.
\newblock Imagenet: A large-scale hierarchical image database.
\newblock In {\em CVPR}. Ieee, 2009.

\bibitem{deng2019arcface}
Jiankang Deng, Jia Guo, Niannan Xue, and Stefanos Zafeiriou.
\newblock {ArcFace}: Additive angular margin loss for deep face recognition.
\newblock In {\em CVPR}, 2019.

\bibitem{deng2020disentangled}
Yu Deng, Jiaolong Yang, Dong Chen, Fang Wen, and Xin Tong.
\newblock Disentangled and controllable face image generation via 3{D}
  imitative-contrastive learning.
\newblock In {\em CVPR}, 2020.

\bibitem{elfwing2018sigmoid}
Stefan Elfwing, Eiji Uchibe, and Kenji Doya.
\newblock Sigmoid-weighted linear units for neural network function
  approximation in reinforcement learning.
\newblock {\em Neural Networks}, 107, 2018.

\bibitem{engelsma2022printsgan}
Joshua~J Engelsma, Steven~A Grosz, and Anil~K Jain.
\newblock Printsgan: synthetic fingerprint generator.
\newblock {\em TPAMI}, 2022.

\bibitem{gecer2018semi}
Baris Gecer, Binod Bhattarai, Josef Kittler, and Tae-Kyun Kim.
\newblock Semi-supervised adversarial learning to generate photorealistic face
  images of new identities from {3D} morphable model.
\newblock In {\em ECCV}, 2018.

\bibitem{geng20193d}
Zhenglin Geng, Chen Cao, and Sergey Tulyakov.
\newblock 3{D} guided fine-grained face manipulation.
\newblock In {\em CVPR}, 2019.

\bibitem{girish2021towards}
Sharath Girish, Saksham Suri, Sai~Saketh Rambhatla, and Abhinav Shrivastava.
\newblock Towards discovery and attribution of open-world gan generated images.
\newblock In {\em ICCV}, 2021.

\bibitem{goodfellow2020generative}
Ian Goodfellow, Jean Pouget-Abadie, Mehdi Mirza, Bing Xu, David Warde-Farley,
  Sherjil Ozair, Aaron Courville, and Yoshua Bengio.
\newblock Generative adversarial networks.
\newblock {\em Communications of the ACM}, 63(11), 2020.

\bibitem{msceleb}
Yandong Guo, Lei Zhang, Yuxiao Hu, Xiaodong He, and Jianfeng Gao.
\newblock {MS-Celeb-1M}: A dataset and benchmark for large-scale face
  recognition.
\newblock In {\em ECCV}, 2016.

\bibitem{he2016deep}
Kaiming He, Xiangyu Zhang, Shaoqing Ren, and Jian Sun.
\newblock Deep residual learning for image recognition.
\newblock In {\em CVPR}, 2016.

\bibitem{fid}
Martin Heusel, Hubert Ramsauer, Thomas Unterthiner, Bernhard Nessler, and Sepp
  Hochreiter.
\newblock Gans trained by a two time-scale update rule converge to a local nash
  equilibrium.
\newblock {\em NeurIPS}, 30, 2017.

\bibitem{ho2020denoising}
Jonathan Ho, Ajay Jain, and Pieter Abbeel.
\newblock Denoising diffusion probabilistic models.
\newblock {\em NeurIPS}, 33, 2020.

\bibitem{ho2022classifier}
Jonathan Ho and Tim Salimans.
\newblock Classifier-free diffusion guidance.
\newblock {\em arXiv preprint arXiv:2207.12598}, 2022.

\bibitem{hu2018disentangling}
Qiyang Hu, Attila Szab{\'o}, Tiziano Portenier, Paolo Favaro, and Matthias
  Zwicker.
\newblock Disentangling factors of variation by mixing them.
\newblock In {\em CVPR}, 2018.

\bibitem{hu2021sail}
Yuan-Ting Hu, Jiahong Wang, Raymond~A Yeh, and Alexander~G Schwing.
\newblock Sail-vos 3d: A synthetic dataset and baselines for object detection
  and 3d mesh reconstruction from video data.
\newblock In {\em CVPR}, 2021.

\bibitem{casia}
Gary Huang, Marwan Mattar, Honglak Lee, and Erik Learned-Miller.
\newblock Learning to align from scratch.
\newblock {\em NeurIPS}, 25, 2012.

\bibitem{lfw}
Gary~B Huang, Marwan Mattar, Tamara Berg, and Eric Learned-Miller.
\newblock Labeled {Faces} in the {Wild}: A database forstudying face
  recognition in unconstrained environments.
\newblock In {\em Workshop on Faces in'Real-Life'Images: Detection, Alignment,
  and Recognition}, 2008.

\bibitem{huang2020curricularface}
Yuge Huang, Yuhan Wang, Ying Tai, Xiaoming Liu, Pengcheng Shen, Shaoxin Li,
  Jilin Li, and Feiyue Huang.
\newblock {CurricularFace}: adaptive curriculum learning loss for deep face
  recognition.
\newblock In {\em CVPR}, 2020.

\bibitem{ioffe2015batch}
Sergey Ioffe and Christian Szegedy.
\newblock Batch normalization: Accelerating deep network training by reducing
  internal covariate shift.
\newblock In {\em ICML}, 2015.

\bibitem{glasses}
Xiaoyi Jiang, Michael Binkert, Bernard Achermann, and Horst Bunke.
\newblock Towards detection of glasses in facial images.
\newblock {\em Pattern Analysis \& Applications}, 3(1), 2000.

\bibitem{ijbs}
Nathan~D Kalka, Brianna Maze, James~A Duncan, Kevin O’Connor, Stephen
  Elliott, Kaleb Hebert, Julia Bryan, and Anil~K Jain.
\newblock {IJB--S}: {IARPA} {Janus} {Surveillance} {Video} {Benchmark}.
\newblock In {\em BTAS}, 2018.

\bibitem{karras2017progressive}
Tero Karras, Timo Aila, Samuli Laine, and Jaakko Lehtinen.
\newblock Progressive growing of gans for improved quality, stability, and
  variation.
\newblock In {\em ICLR}, 2018.

\bibitem{karras2019style}
Tero Karras, Samuli Laine, and Timo Aila.
\newblock A style-based generator architecture for generative adversarial
  networks.
\newblock In {\em CVPR}, 2019.

\bibitem{karras2020analyzing}
Tero Karras, Samuli Laine, Miika Aittala, Janne Hellsten, Jaakko Lehtinen, and
  Timo Aila.
\newblock Analyzing and improving the image quality of stylegan.
\newblock In {\em CVPR}, 2020.

\bibitem{kim2018deep}
Hyeongwoo Kim, Pablo Garrido, Ayush Tewari, Weipeng Xu, Justus Thies, Matthias
  Niessner, Patrick P{\'e}rez, Christian Richardt, Michael Zollh{\"o}fer, and
  Christian Theobalt.
\newblock Deep video portraits.
\newblock {\em TOG}, 2018.

\bibitem{kim2022adaface}
Minchul Kim, Anil~K Jain, and Xiaoming Liu.
\newblock {AdaFace}: Quality adaptive margin for face recognition.
\newblock In {\em CVPR}, 2022.

\bibitem{caface}
Minchul Kim, Feng Liu, Anil Jain, and Xiaoming Liu.
\newblock Cluster and aggregate: Face recognition with large probe set.
\newblock {\em NeurIPS}, 2022.

\bibitem{kingma2014adam}
Diederik~P Kingma and Jimmy Ba.
\newblock Adam: A method for stochastic optimization.
\newblock In {\em ICLR}, 2015.

\bibitem{kupas2021solving}
David Kupas and Balazs Harangi.
\newblock Solving the problem of imbalanced dataset with synthetic image
  generation for cell classification using deep learning.
\newblock In {\em EMBC}, 2021.

\bibitem{kynkaanniemi2019improved}
Tuomas Kynk{\"a}{\"a}nniemi, Tero Karras, Samuli Laine, Jaakko Lehtinen, and
  Timo Aila.
\newblock Improved precision and recall metric for assessing generative models.
\newblock {\em NeurIPS}, 32, 2019.

\bibitem{lee2019srm}
HyunJae Lee, Hyo-Eun Kim, and Hyeonseob Nam.
\newblock Srm: A style-based recalibration module for convolutional neural
  networks.
\newblock In {\em ICCV}, 2019.

\bibitem{lin2018conditional}
Jianxin Lin, Yingce Xia, Tao Qin, Zhibo Chen, and Tie-Yan Liu.
\newblock Conditional image-to-image translation.
\newblock In {\em CVPR}, 2018.

\bibitem{controllable-and-guided-face-synthesis-for-unconstrained-face-recognition}
Feng Liu, Minchul Kim, Anil Jain, and Xiaoming Liu.
\newblock Controllable and guided face synthesis for unconstrained face
  recognition.
\newblock In {\em ECCV}, 2022.

\bibitem{liu2017sphereface}
Weiyang Liu, Yandong Wen, Zhiding Yu, Ming Li, Bhiksha Raj, and Le Song.
\newblock {SphereFace}: Deep hypersphere embedding for face recognition.
\newblock In {\em CVPR}, 2017.

\bibitem{9779478}
Yaojie Liu and Xiaoming Liu.
\newblock Spoof trace disentanglement for generic face anti-spoofing.
\newblock {\em TPAMI}, 45(3), 2023.

\bibitem{loshchilov2017decoupled}
Ilya Loshchilov and Frank Hutter.
\newblock Decoupled weight decay regularization.
\newblock {\em arXiv preprint arXiv:1711.05101}, 2017.

\bibitem{most-gan-3d-morphable-stylegan-for-disentangled-face-image-manipulation}
Safa~C. Medin, Bernhard Egger, Anoop Cherian, Ye Wang, Joshua~B. Tenenbaum,
  Xiaoming Liu, and Tim~K. Marks.
\newblock {MOST-GAN}: 3d morphable stylegan for disentangled face image
  manipulation.
\newblock In {\em AAAI}, 2022.

\bibitem{agedb}
Stylianos Moschoglou, Athanasios Papaioannou, Christos Sagonas, Jiankang Deng,
  Irene Kotsia, and Stefanos Zafeiriou.
\newblock {AGEDB}: the first manually collected, in-the-wild age database.
\newblock In {\em CVPRW}, 2017.

\bibitem{nguyen2019hologan}
Thu Nguyen-Phuoc, Chuan Li, Lucas Theis, Christian Richardt, and Yong-Liang
  Yang.
\newblock Holo{GAN}: Unsupervised learning of 3d representations from natural
  images.
\newblock In {\em ICCV}, 2019.

\bibitem{nichol2021improved}
Alexander~Quinn Nichol and Prafulla Dhariwal.
\newblock Improved denoising diffusion probabilistic models.
\newblock In {\em ICML}, pages 8162--8171. PMLR, 2021.

\bibitem{piao2019semi}
Jingtan Piao, Chen Qian, and Hongsheng Li.
\newblock Semi-supervised monocular 3{D} face reconstruction with end-to-end
  shape-preserved domain transfer.
\newblock In {\em ICCV}, 2019.

\bibitem{diffae}
Konpat Preechakul, Nattanat Chatthee, Suttisak Wizadwongsa, and Supasorn
  Suwajanakorn.
\newblock Diffusion autoencoders: Toward a meaningful and decodable
  representation.
\newblock In {\em CVPR}, 2022.

\bibitem{pumarola2018ganimation}
Albert Pumarola, Antonio Agudo, Aleix~M Martinez, Alberto Sanfeliu, and
  Francesc Moreno-Noguer.
\newblock Ganimation: Anatomically-aware facial animation from a single image.
\newblock In {\em ECCV}, 2018.

\bibitem{qiu2021synface}
Haibo Qiu, Baosheng Yu, Dihong Gong, Zhifeng Li, Wei Liu, and Dacheng Tao.
\newblock {SynFace}: Face recognition with synthetic data.
\newblock In {\em ICCV}, 2021.

\bibitem{dalle2}
Aditya Ramesh, Prafulla Dhariwal, Alex Nichol, Casey Chu, and Mark Chen.
\newblock Hierarchical text-conditional image generation with clip latents.
\newblock {\em arXiv preprint arXiv:2204.06125}, 2022.

\bibitem{rombach2022high}
Robin Rombach, Andreas Blattmann, Dominik Lorenz, Patrick Esser, and Bj{\"o}rn
  Ommer.
\newblock High-resolution image synthesis with latent diffusion models.
\newblock In {\em CVPR}, 2022.

\bibitem{salimans2016improved}
Tim Salimans, Ian Goodfellow, Wojciech Zaremba, Vicki Cheung, Alec Radford, and
  Xi Chen.
\newblock Improved techniques for training gans.
\newblock {\em NeurIPS}, 29, 2016.

\bibitem{cfpfp}
Soumyadip Sengupta, Jun-Cheng Chen, Carlos Castillo, Vishal~M Patel, Rama
  Chellappa, and David~W Jacobs.
\newblock Frontal to profile face verification in the wild.
\newblock In {\em WACV}, 2016.

\bibitem{sha2022fake}
Zeyang Sha, Zheng Li, Ning Yu, and Yang Zhang.
\newblock De-fake: Detection and attribution of fake images generated by
  text-to-image diffusion models.
\newblock {\em arXiv preprint arXiv:2210.06998}, 2022.

\bibitem{shen2018facefeat}
Yujun Shen, Bolei Zhou, Ping Luo, and Xiaoou Tang.
\newblock Facefeat-{GAN}: a two-stage approach for identity-preserving face
  synthesis.
\newblock {\em arXiv preprint arXiv:1812.01288}, 2018.

\bibitem{sohl2015deep}
Jascha Sohl-Dickstein, Eric Weiss, Niru Maheswaranathan, and Surya Ganguli.
\newblock Deep unsupervised learning using nonequilibrium thermodynamics.
\newblock In {\em ICML}, 2015.

\bibitem{song2020denoising}
Jiaming Song, Chenlin Meng, and Stefano Ermon.
\newblock Denoising diffusion implicit models.
\newblock In {\em ICLR}, 2021.

\bibitem{song2021maximum}
Yang Song, Conor Durkan, Iain Murray, and Stefano Ermon.
\newblock Maximum likelihood training of score-based diffusion models.
\newblock {\em NeurIPS}, 34, 2021.

\bibitem{song2019generative}
Yang Song and Stefano Ermon.
\newblock Generative modeling by estimating gradients of the data distribution.
\newblock {\em NeurIPS}, 32, 2019.

\bibitem{song2020improved}
Yang Song and Stefano Ermon.
\newblock Improved techniques for training score-based generative models.
\newblock {\em NeurIPS}, 33:12438--12448, 2020.

\bibitem{noise-modeling-synthesis-and-classification-for-generic-object-anti-spoofing}
Joel Stehouwer, Amin Jourabloo, Yaojie Liu, and Xiaoming Liu.
\newblock Noise modeling, synthesis and classification for generic object
  anti-spoofing.
\newblock In {\em CVPR}, 2020.

\bibitem{sun2019single}
Tiancheng Sun, Jonathan~T Barron, Yun-Ta Tsai, Zexiang Xu, Xueming Yu, Graham
  Fyffe, Christoph Rhemann, Jay Busch, Paul~E Debevec, and Ravi Ramamoorthi.
\newblock Single image portrait relighting.
\newblock {\em TOG}, 2019.

\bibitem{disentangled-representation-learning-gan-for-pose-invariant-face-recognition}
Luan Tran, Xi Yin, and Xiaoming Liu.
\newblock Disentangled representation learning gan for pose-invariant face
  recognition.
\newblock In {\em CVPR}, 2017.

\bibitem{tremblay2018training}
Jonathan Tremblay, Aayush Prakash, David Acuna, Mark Brophy, Varun Jampani, Cem
  Anil, Thang To, Eric Cameracci, Shaad Boochoon, and Stan Birchfield.
\newblock Training deep networks with synthetic data: Bridging the reality gap
  by domain randomization.
\newblock In {\em CVPRW}, 2018.

\bibitem{van2021decaf}
Boris van Breugel, Trent Kyono, Jeroen Berrevoets, and Mihaela van~der Schaar.
\newblock Decaf: Generating fair synthetic data using causally-aware generative
  networks.
\newblock {\em NeurIPS}, 34:22221--22233, 2021.

\bibitem{van2008visualizing}
Laurens Van~der Maaten and Geoffrey Hinton.
\newblock Visualizing data using {t-SNE}.
\newblock {\em Journal of Machine Learning Research}, 2008.

\bibitem{vaswani2017attention}
Ashish Vaswani, Noam Shazeer, Niki Parmar, Jakob Uszkoreit, Llion Jones,
  Aidan~N Gomez, {\L}ukasz Kaiser, and Illia Polosukhin.
\newblock Attention is all you need.
\newblock In {\em NeurIPS}, 2017.

\bibitem{wang2018cosface}
Hao Wang, Yitong Wang, Zheng Zhou, Xing Ji, Dihong Gong, Jingchao Zhou, Zhifeng
  Li, and Wei Liu.
\newblock {CosFace}: Large margin cosine loss for deep face recognition.
\newblock In {\em CVPR}, 2018.

\bibitem{wang2020cnn}
Sheng-Yu Wang, Oliver Wang, Richard Zhang, Andrew Owens, and Alexei~A Efros.
\newblock Cnn-generated images are surprisingly easy to spot... for now.
\newblock In {\em CVPR}, 2020.

\bibitem{piti}
Tengfei Wang, Ting Zhang, Bo Zhang, Hao Ouyang, Dong Chen, Qifeng Chen, and
  Fang Wen.
\newblock Pretraining is all you need for image-to-image translation.
\newblock {\em arXiv preprint arXiv:2205.12952}, 2022.

\bibitem{ijbb}
Cameron Whitelam, Emma Taborsky, Austin Blanton, Brianna Maze, Jocelyn Adams,
  Tim Miller, Nathan Kalka, Anil~K Jain, James~A Duncan, Kristen Allen, et~al.
\newblock {IARPA} {Janus} {Benchmark}-{B} face dataset.
\newblock In {\em CVPRW}, 2017.

\bibitem{glassesgit}
Tianxing Wu.
\newblock Realtime glasses detection.
\newblock \url{https://github.com/TianxingWu/realtime-glasses-detection}, 2022.

\bibitem{wu2018group}
Yuxin Wu and Kaiming He.
\newblock Group normalization.
\newblock In {\em ECCV}, 2018.

\bibitem{wyzykowski2022synthetic}
Andre Brasil~Vieira Wyzykowski and Anil~K Jain.
\newblock Synthetic latent fingerprint generator.
\newblock In {\em WACV}, 2023.

\bibitem{xiao2018elegant}
Taihong Xiao, Jiapeng Hong, and Jinwen Ma.
\newblock Elegant: Exchanging latent encodings with {GAN} for transferring
  multiple face attributes.
\newblock In {\em ECCV}, 2018.

\bibitem{yi2014learning}
Dong Yi, Zhen Lei, Shengcai Liao, and Stan~Z Li.
\newblock Learning face representation from scratch.
\newblock {\em arXiv preprint arXiv:1411.7923}, 2014.

\bibitem{yu2019attributing}
Ning Yu, Larry~S Davis, and Mario Fritz.
\newblock Attributing fake images to gans: Learning and analyzing gan
  fingerprints.
\newblock In {\em ICCV}, 2019.

\bibitem{zhang2016joint}
Kaipeng Zhang, Zhanpeng Zhang, Zhifeng Li, and Yu Qiao.
\newblock Joint face detection and alignment using multitask cascaded
  convolutional networks.
\newblock {\em Signal Processing Letters}, 2016.

\bibitem{cplfw}
Tianyue Zheng and Weihong Deng.
\newblock Cross-{Pose} {LFW}: A database for studying cross-pose face
  recognition in unconstrained environments.
\newblock {\em Beijing University of Posts and Telecommunications, Tech. Rep},
  5, 2018.

\bibitem{calfw}
Tianyue Zheng, Weihong Deng, and Jiani Hu.
\newblock Cross-{Age} {LFW:} {A} database for studying cross-age face
  recognition in unconstrained environments.
\newblock {\em CoRR}, abs/1708.08197, 2017.

\bibitem{zhu2021webface260m}
Zheng Zhu, Guan Huang, Jiankang Deng, Yun Ye, Junjie Huang, Xinze Chen, Jiagang
  Zhu, Tian Yang, Jiwen Lu, Dalong Du, et~al.
\newblock {WebFace260M}: A benchmark unveiling the power of million-scale deep
  face recognition.
\newblock In {\em CVPR}, 2021.

\bibitem{zunair2021synthesis}
Hasib Zunair and A~Ben Hamza.
\newblock Synthesis of covid-19 chest x-rays using unpaired image-to-image
  translation.
\newblock {\em Social network analysis and mining}, 11(1), 2021.

\end{thebibliography}
}

\onecolumn
\setcounter{equation}{0}
\setcounter{figure}{0}
\setcounter{table}{0}
\setcounter{page}{1}
\setcounter{section}{0}

\begin{center}
\textbf{\Large DCFace: Synthetic Face Generation with Dual Condition Diffusion Model}\\
\vspace{2mm}
\textbf{\large Supplementary Material}\\
\end{center}

\colorlet{cblack}{black!40!black}
\colorlet{cblue}{blue!40!gray}
\colorlet{cred}{red!40!gray}
\colorlet{cyellow}{yellow!40!gray}
\colorlet{cgreen}{green!40!gray}
\colorlet{ccyan}{cyan!40!gray}
\colorlet{cpurple}{purple!40!gray}

\newcommand{\Esty}{\textcolor{cgreen}{$E_{sty}$}}
\newcommand{\LID}{\textcolor{cgreen}{$L_{ID}$}}
\newcommand{\cellgray}{\cellcolor[HTML]{e5e5e5}}

\renewcommand\thesection{\Alph{section}}

\section{Training Details}
\subsection{Architecture Detals}

\begin{figure}[h!]
    \centering
    \includegraphics[width=\linewidth]{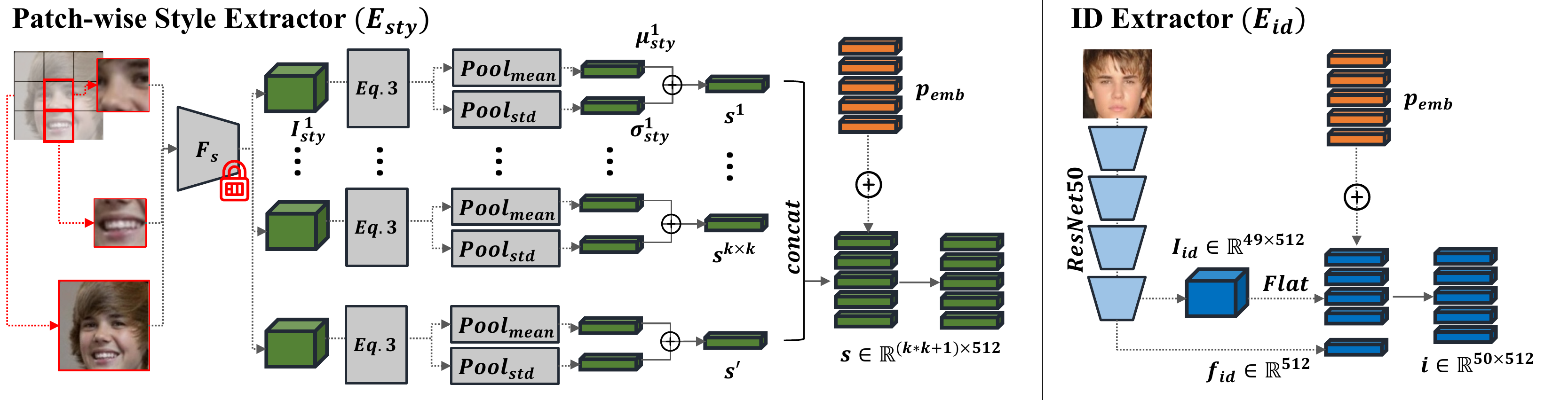}
    \caption{Left: An illustration of $\bm{X}_{sty}$. The key property of $\bm{X}_{sty}$ is in restricting the information in $\bm{X}_{sty}$ from flowing freely to the next layer. The fixed feature encoder $\bm{F}_s$ and the patch-wise spatial mean-variance operation destroy the detailed ID information while preserving the style of an image. We create an output of size $\mathbb{R}^{(k^2+1)\!\times\!C}$. Right: A simple CNN based on ResNet50. We take intermediate representation and the last feature vector and concatenate them together to create a output of size $\mathbb{R}^{50\!\times\!C}$. }
    \label{fig:supp1}
\end{figure}

\begin{figure}
    \centering
    \includegraphics[width=\linewidth]{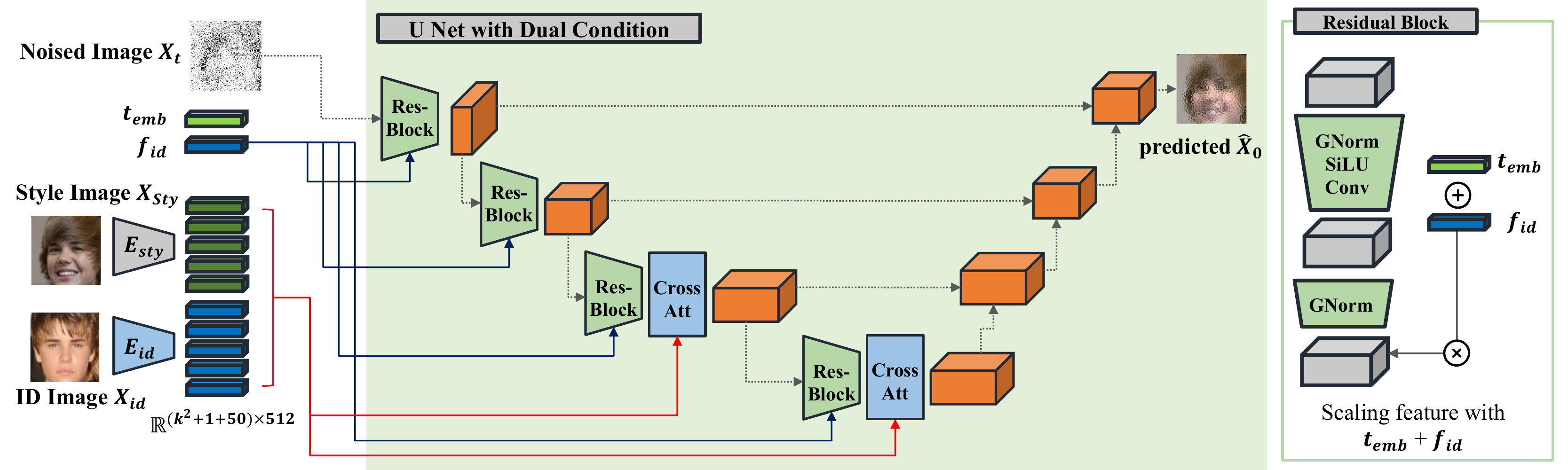}
    \caption{Illustration of DDPM U-Net with conditioning operations highlighted. The red arrow indicates how the dual conditions are injected into the intermediate features of U-Net using cross-attention layers. For clarity, up-sampling stages are not illustrated, but they are symmetric to the down-sampling stages. On the right is a detailed illustration of the Residual Block with timestep and ID condition. $\bm{t}_{emb}$ and $\bm{f}_{id}$ from $E_{id}$ are added together and used to scale the output of the Residual Block. }
    \label{fig:supp2}
\end{figure}

The dual condition generator $G_{mix}$ is a modification of DDPM~\cite{ho2020denoising} to incorporate two conditions. We insert two conditions $\bm{X}_{id}$ and $\bm{X}_{sty}$ into the denoising U-Net $\bm{\epsilon}_\theta(\bm{X}_{t},t,\bm{X}_{id}, \bm{X}_{sty})$. Conditioning images $\bm{X}_{sty}$ and $\bm{X}_{id}$ are mapped to features using $E_{sty}$ and $E_{id}$, respectively. According to Eq.~6 of the main paper, the style information $E_{sty}(\bm{X}_{sty}) $ is the concatenation of style vectors at different $k\!\times\!k$ patch locations, 
\begin{equation}
E_{sty}(\bm{X}_{sty}) := \bm{s} = \left[\bm{s}^1, \bm{s}^2, \bm{s}^{k_i}..., \bm{s}^{k\!\times\!k}, \bm{s}'\right] \in \mathbb{R}^{(k^2+1)\!\times\!C}.
\end{equation}
On the other hand, ID information is a concatenation of features extracted from a trainable CNN (e.g. ResNet50~\cite{he2016deep}), which produces an intermediate feature $\bm{I}_{id}$ of shape $\mathbb{R}^{7\!\times\!7\!\times\!512}$and a feature vector $\bm{f}_{id}$ of shape $\mathbb{R}^{512}$. Specifically, 
\begin{align}
    E_{id}(\bm{X}_{id}) := \bm{i} = \left[ \text{Flatten}(\bm{I}_{id}), \bm{f}_{id}  \right] + \bm{P}_{emb} \in \mathbb{R}^{50\!\times\!C}, 
\end{align}
where Flatten refers to removing the $H\!\times\!W$ spatial dimension and $\mathbb{R}^{50\!\times\!C}$ is from concatenating features of length $7\!*\!7$ and $1$. $\bm{P}_{emb}$ is a learnable position embedding for distinguishing each feature position for the subsequent cross-attention operation. Detailed illustrations of 
$E_{sty}(\bm{X}_{sty})$ and $E_{id}(\bm{X}_{id})$ are shown in Fig.~\ref{fig:supp1}. $C$ for the channel dimension of $E_{sty}(\bm{X}_{sty}) $ and $E_{id}(\bm{X}_{id}) $ is $512$. 
\clearpage
When $E_{sty}(\bm{X}_{sty})$ and $E_{id}(\bm{X}_{id})$ is prepared, they together form $(k^2+1) + 50$ vectors of shape $512$. These can be injected into the U-Net $\bm{\epsilon}_\theta$ by following the convention of the DDPM based text-conditional image generators~\cite{dalle2}. Specifically, cross attention operation can be written as 
a modification of attention equation~\cite{vaswani2017attention} with query $\bm{Q}$, key $\bm{K}$ and value $\bm{V}$ with additional query $\bm{Q}_c$, key $\bm{K}_c$.  
\begin{align}
    \text{Attn}(\bm{Q}, \bm{K}, \bm{V})&=\text{SoftMax}\left(  \frac{\bm{Q}\bm{W}_q \left(\bm{K} \bm{W}_k \right)^\intercal}{\sqrt{d}} \right) \bm{W}_v \bm{V},\\
    \text{Cross-Attn}(\bm{Q}, \bm{K}, \bm{V},  \bm{K}_c, \bm{V}_c)&=\text{SoftMax}\left(  \frac{\bm{Q}\bm{W}_q \left([\bm{K}, \bm{K}_c] \bm{W}_k \right)^\intercal}{\sqrt{d}} \right) \bm{W}_v [\bm{V}, \bm{V}_c],
\end{align}
where $\bm{W}_q, \bm{W}_k$ and $\bm{W}_v$ are learnable weights and $[\cdot]$ refers to concatenation operation. In our case, $\bm{Q}\!=\!\bm{K}\!=\!\bm{V}$ are an arbitrary intermediate feature in the U-Net. And $\bm{K}_c=\bm{V}_c$ are conditions generated by $E_{sty}(\bm{X}_{sty})$ and $E_{id}(\bm{X}_{id})$, concatenated together. This operation allows the model to update the intermediate features with the conditions if necessary. We insert the cross-attention module in the last two DownSampling Residual Blocks in the U-Net, as shown in Fig.~\ref{fig:supp2}. 

To increase the effect of $\bm{X}_{id}$ in the conditioning operation, we also add $\bm{f}_{id}$ to the time-step embedding $\bm{t}_{emb}$. As shown in the right side of Fig.~\ref{fig:supp2}, the Residual Block in the U-Net modulates the intermediate features according to the scaling vector provided by $\bm{f}_{id} + \bm{t}_{emb}$. GNorm~\cite{wu2018group} refers to Group Normalization and SiLU refers to Sigmoid Linear Units~\cite{elfwing2018sigmoid}. Adding $\bm{f}_{id}$ to $\bm{t}_{emb}$ for the Residual Block allows more paths for $\bm{X}_{id}$ to change the output of U-Net. 

\subsection{Training Hyper-Parameters}
The final loss for training the model end-to-end is $L_{MSE}+\lambda L_{ID}$ with $\lambda$ as a scaling parameter. We set $\lambda=0.05$ to compensate for the different scale between L2 and Cosine Similarity. All our input image sizes are $112\!\times\!112$, following the convention of SoTA face recognition model datasets~\cite{casia,zhu2021webface260m,deng2019arcface}. And our code is implemented in Pytorch. 

\clearpage
\section{More Experiment Results}
\subsection{Adding Real Dataset}
We include additional experiment results that involve adding real images. Although the motivation of the paper is to use an only-synthetic dataset to train a face recognition model, the performance comparison with an addition of a subset of the real dataset has its merits; it shows 1) whether the synthetic dataset is complementary to the real dataset and 2) whether the synthetic dataset can work as an augmentation for real images.   

Tab.~\ref{tab:realadd} shows the performance comparison between DigiFace~\cite{bae2022digiface} and our proposed DCFace when 1) a few real images are added and 2) both synthetic datasets are combined. The performance gap for DigiFace is large, jumping from $86.37$ to $92.67$ on average when $2K$ real subjects with $20$ images per subject are added. In contrast, ours show a relatively less dramatic gain, $91.21$ to $92.90$ when few real images are added. This indicates that DigiFace~\cite{bae2022digiface} is quite different from the real images and ours is similar to the real images. This is in-line with our expectation as we have created a synthetic dataset that tries to mimic the style distribution of the training dataset, whereas DigiFace simulates image styles using 3D models. 

\subsection{Combining Multiple Synthetic Datasets}

In the second to the last row of Tab.~\ref{tab:realadd}, when we combined the two synthetic datasets without the real images, the performance is the highest, reaching $93.06$ on average. This result indicates that different synthetic datasets can be complementary when they are generated using different methods. 

\definecolor{gray}{rgb}{0.87, 0.87, 0.87}
\begin{table}[h]
\centering
\small
\begin{tabular}{|c|c|c|c|c|c|c|c||c|c|}
\hline
         & \# Synthetic   Imgs       & \# Real Imgs & LFW   & CFPFP & CPLFW & AGEDB & CALFW & AVG  &\fillg  \makecell{Gap to\\  Real}  \\ \hline
DigiFace & $1.2$M ($10\text{K}\!\times\!72\!+\!100\text{K}\!\times\!5$)  & 0              & $96.17$ & $89.81$ & $82.23$ & $81.10$ & $82.55$ & $86.37$ & \fillg $8.72$ \\ 
DigiFace & $1.2$M ($10\text{K}\!\times\!72\!+\!100\text{K}\!\times\!5$)   & 2K×20          & $99.17$ & $94.63$ & $88.1$  & $90.5$  & $90.97$ & $92.67$ & \fillg $2.06$\\ \hline\hline
DCFace   & $1.2$M ($20\text{K}\!\times\!50\!+\!40\text{K}\!\times\!5$)  & 0              & $98.58$ & $88.61$ & $85.07$ & $90.97$ & $92.82$ & $91.21$ & \fillg $3.61$ \\ 
DCFace   & 1.2M ($20\text{K}\!\times\!50\!+\!40\text{K}\!\times\!5$) & 2K×20          & $98.97$ & $94.01$ & $86.78$ & $91.80$ & $92.95$ & $92.90$ & \fillg  $1.82$\\ \hline\hline
\multicolumn{2}{|c|}{DCFace+DigiFace (2.4M) }   & 0              & $99.20$ & $93.63$ & $87.25$ & $92.25$ & $92.95$ & $93.06$ & \fillg  $\bm{1.65}$\\ \hline\hline
CASIA   & 0   & 0.5M            &  $99.42$ &  $96.56$ &  $89.73$  & $94.08$  & $93.32$ &  $94.62$ & \fillg  $0$ \\ \hline
\end{tabular}
\caption{Verification accuracies of FR models trained with synthetic datasets and subset of real datasets. In all settings, the backbone is set to IR50~\cite{deng2019arcface} model with AdaFace loss~\cite{kim2022adaface} for a fair comparison. }
\label{tab:realadd}
\end{table}

\clearpage
\section{Analysis}
\begin{figure}[h]
    \centering
    \includegraphics[width=0.47\linewidth]{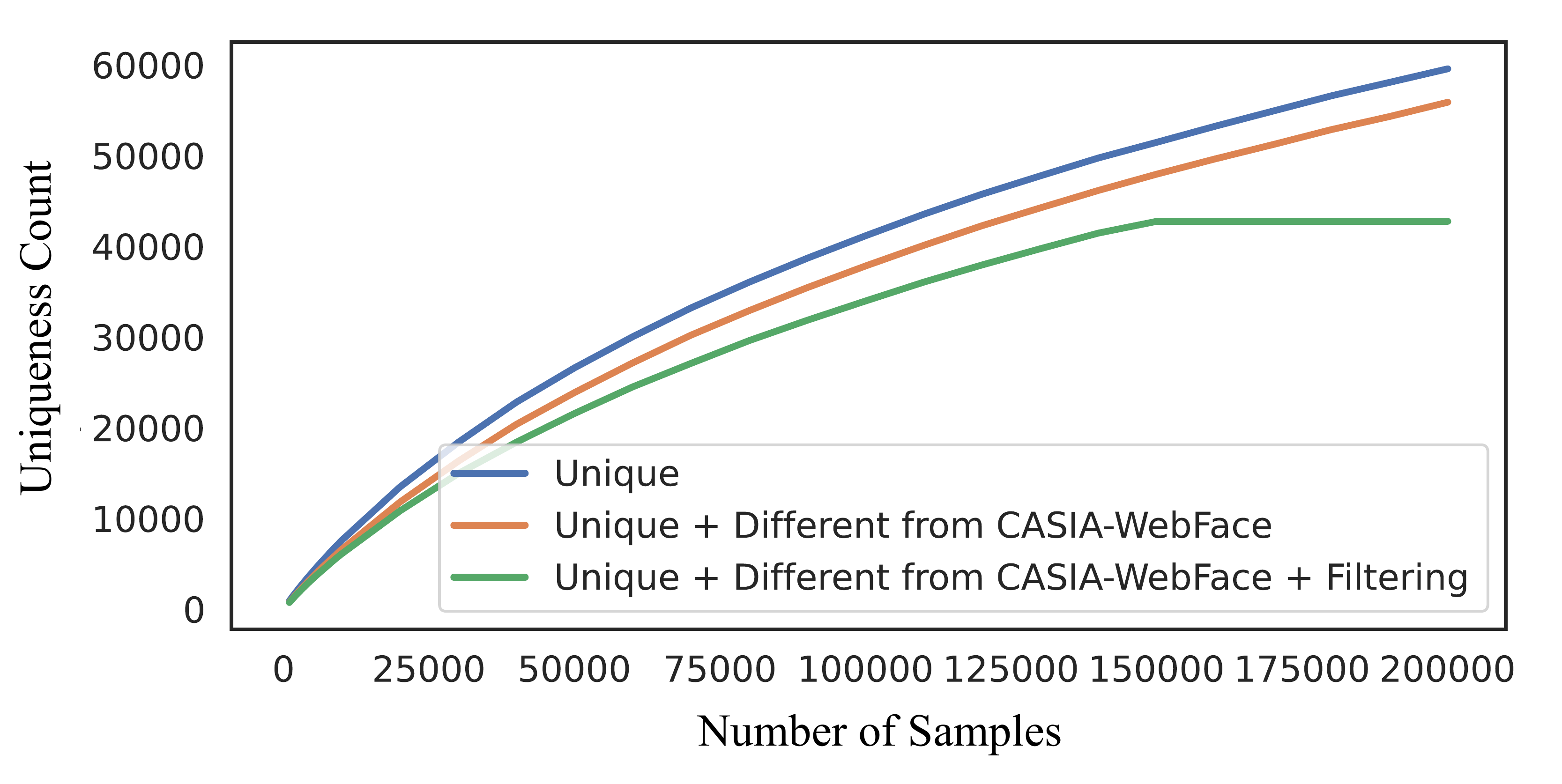}
    \vspace{-3mm}
    \caption{Plot of unique subject count as the number of samples from $G_{id}$ is increased from $1000$ to $200,000$. At $200,000$, one additional sample has approximately $15\%$ chance of being unique. And the rate decreases with more samples. }
    \label{fig:uniquenesscount}
    \vspace{-2mm}
\end{figure}

\Paragraph{C.1 Unique Subject Counts}
In Fig.~\ref{fig:uniquenesscount}, we plot the number of unique subjects that can be sampled as we increase the sample size. The blue curve shows that the number of unique samples that can be generated by a DDPM of our choice does not saturate when we sample $200,000$ samples. At $200,000$ samples, the unique subjects are about $60,000$. And by extrapolating the curve, we estimate the number might reach $80,000$ with more samples. Our DDPM of choice is trained on FFHQ~\cite{karras2019style} dataset which contains $70,000$ unlabeled high-quality images. The orange line shows the number of unique samples that are sufficiently different from the subjects in the CASIA-WebFace dataset. The green line shows the number of unique samples left after filtering images that contain sunglasses. The flat region is due to the filtering stage reducing the total candidates. The plot shows that DDPM trained on FFHQ dataset can sufficiently generate a large number of unique and new samples that are different from CASIA-WebFace dataset. However, with more samples, eventually there is a limit to the number of unique samples that can be generated. When the number of total generated samples is $100,000$, one additional sample has approximately $24\%$ chance of being unique, whereas, at $200,000$, the probability is $15\%$. The rate of sampling another unique subject decreases with more samples. The model used for evaluating the uniqueness is IR101~\cite{deng2019arcface} trained on the WebFace4M~\cite{zhu2021webface260m} dataset. And we use the threshold of $0.3$. 
We would like to note a typo in Sec.~3.3 of the main paper, where the number of unique subjects should be corrected from $62,570$ to $42,763$. 
   \vspace{-3mm}

\begin{figure}[h]
    \centering
    \includegraphics[width=\linewidth]{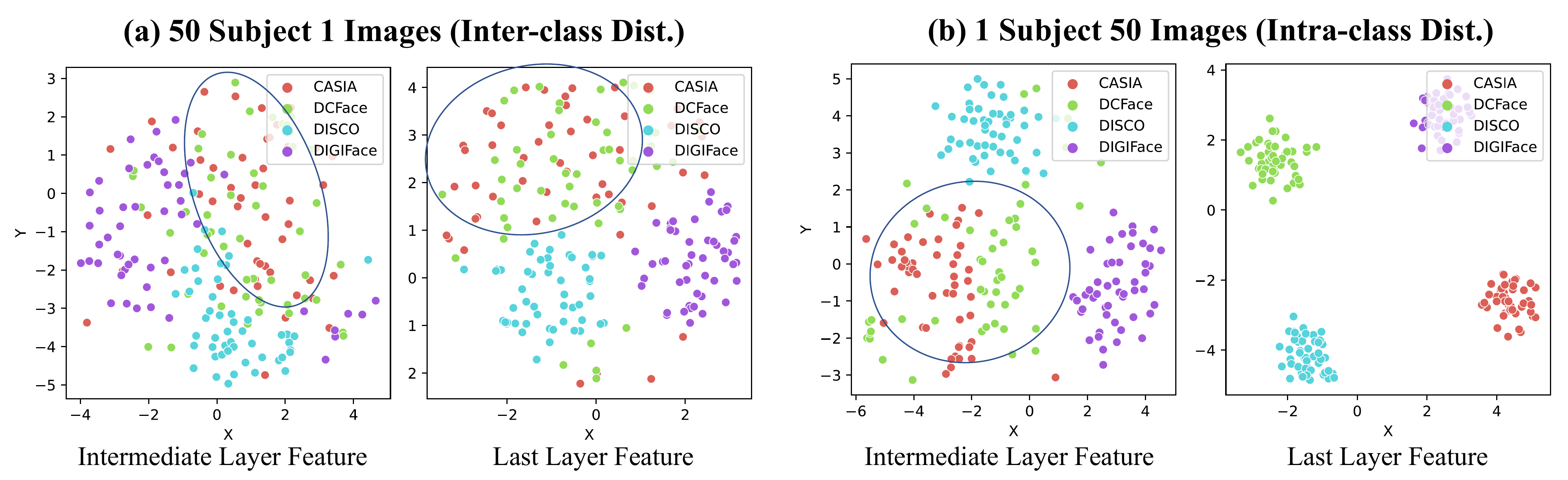}
    \vspace{-3mm}
    \caption{(a) the t-SNE plot of features from synthetic and real datasets of $50$ subjects per dataset. It shows how $50$ randomly sampled subjects from each dataset are distributed. The distribution between real (red) and DCFace (green) is the closest. (b)  the t-SNE plot of features from synthetic and real datasets of $1$ subject per dataset with $50$ images. We randomly sample $1$ subject from each dataset. The last layer features are well separated as the model is a face recognition model that separates the features of different subjects. }
    \label{fig:featureComp}
\end{figure}

\Paragraph{C.2 Feature Plot}
In Fig.~\ref{fig:featureComp}, we show the 2D t-SNE~\cite{van2008visualizing} plot of synthetic images generated by $3$ different methods (DiscoFaceGAN~\cite{deng2020disentangled}, DigiFace~\cite{bae2022digiface} and proposed DCFace). The red circles represent real images from CASIA-WebFace. We extract the features from each image using a pre-trained face recognition model, IR101~\cite{deng2019arcface} trained on WebFace4M~\cite{zhu2021webface260m}. 
We show two settings we sample (a) $50$ subjects with $1$ image per subject and (b) $1$ subject with $50$ images per subject. Note that the proximity of DCFace image features is closer to CASIA-WebFace image features, highlighted in a circle. For each setting, we show the features extracted from an intermediate layer of IR101 and the last layer. As the layer becomes deeper, the features become suitable for recognition, as shown in the last column of the figure. 
\vspace{-3mm}

\Paragraph{C.3 Comparison with Classifier Free Guidance}
 
When $\bm{\epsilon}(x_t, c)$ learns to use the condition $c$, the difference $\bm{\epsilon}(x_t, c)\!-\!\bm{\epsilon}(x_t)$ can give further guidance during sampling to increase the dependence on $c$. But, in our case, the ID condition is the fine-grained facial difference that is hard to learn with MSE loss. Proposed Time-dependent ID loss, \LID~helps the model learn this directly. Row 3 vs 4 of Tab.~\ref{tab:guide} shows that \LID~is more effective than CFG. 

\begin{table}[h]
\centering
\begin{tabular}{|c|c|c|c|c|c|}
\hline
 & \cellcolor[HTML]{FFC702}\textbf{Conditions} & \cellcolor[HTML]{FFC702}\textbf{Train Loss}  & \cellcolor[HTML]{FFC702}\textbf{Sampling} & \cellcolor[HTML]{FFC702}\textbf{FR.Perf} $\uparrow$ \\ \hline\hline
1 & \cellgray CNN($X_{id}$), CNN($X_{sty}$)      & \cellgray MSE    & \cellgray + Guide               &   \cellgray                                        $73.38$       \\ \hline\hline
2 & CNN($X_{id}$), \Esty($X_{sty}$)        & MSE      &    $\times$       &               $82.30$             \\ 
3 & CNN($X_{id}$), \Esty($X_{sty}$)        & MSE      & + Guide          &             $84.05$             \\ 
4 & CNN($X_{id}$), \Esty($X_{sty}$)            & MSE+\LID     &   $\times$    &        $\bm{89.56}$                                                \\ \hline
\end{tabular}
\caption{Green \Esty~and \LID~indicates the novelty of our paper. For guidance, we adopt $10\%$ condition masking during training and the guidance scale of 3 during sampling. FR.Perf is an average of 5 face recognition performances as in the main paper.
}
\label{tab:guide}
\end{table}

Interestingly, with a large guidance scale, CFG becomes harmful. CFG decreases diversity as pointed out by ~\cite{ho2022classifier}. We observe that guidance with $X_{id}$ leads to consistent ID but with little facial variation, the same phenomenon in DCFace with grid-size 1x1 in $E_{sty}$, in Tab.~2 (main). Good FR datasets need both large intra and inter-subject variability and we combine \Esty~and \LID~to achieve this.
\vspace{2mm}

\Paragraph{C.4 FID Scores}
Note that our generated data is not high-res images like FFHQ when compared to how SynFace is similar to FFHQ. 
(Tab.~\ref{tab:fid} row 5 vs 6).
But, we point out that our aim is not to create HQ images but to create a \textit{database} with realistic inter/intra-subject variations. In that regard, we have successfully approximated  the distribution of the popular FR training dataset CASIA-WebFace (FID=13.67). 

\begin{table}[h]
\centering
\setlength{\tabcolsep}{4pt}
\begin{tabular}{|c|c||c|c|c|}
\hline
& \cellgray Generator Train Data & \cellcolor[HTML]{FFC702} Source (real/syn)  & \cellcolor[HTML]{FFC702} Target (real)    & \cellcolor[HTML]{FFC702}FID $\downarrow$  \\ \hline\hline
1 & \cellgray  - & CASIA (train) & CASIA (val)     & $\bm{9.57}$                                    \\ \hline\hline
2 & \cellgray CASIA (train)  & DCFace       & CASIA (val)     & $\bm{13.67}$                           \\ 
3 & \cellgray FFHQ+3DMM    & SynFace & CASIA (val)     & $38.48$                       \\ 
4 & \cellgray 3D Face Capture  &DIGIFACE1M       & CASIA (val)     & $71.65$               \\ \hline\hline
5 & \cellgray CASIA (train)    & DCFace       & FFHQ (train+val) & $35.45$                         \\ 
6& \cellgray FFHQ+3DMM  & SynFace & FFHQ (train+val) & $\bm{21.75}$                         \\ 

7 & \cellgray 3D Face Capture  & DIGIFACE1M       & FFHQ (train+val) & $68.67$               \\ \hline
\end{tabular}

\caption{FID scores of synthetic  vs real datasets. For synthetic datasets, we randomly sampled $10,000$ images. See Line 630 for Casia-WebFace Train and Val set split. All images are aligend and cropped to $112\!\times\!112$ to be in accordance with CASIA-WebFace. }
\label{tab:fid}
\end{table}

Having said this, we note FID is not comprehensive in evaluating labeled datasets. It cannot capture the label consistency nor directly relate to the FR performance. As such, SynFace/DigiFace do not report FID. We propose U,D,C metrics that enable holistic analysis of labeled datasets. 

\vspace{2mm}
\Paragraph{C.5 Does DCFace change gender?} 
DCFace combines $X_{ID}$ and $X_{sty}$, while adhering to the subject ID as defined by a pre-trained FR model. Factors weakly related to ID, such as age and hair style, can vary. Biometric ambiguity can occur due to makeup, wig, weight change, \textit{etc.}~even in real life. The perceived gender may change, but changes such as hair are less relevant to subject ID for the FR model.

\vspace{2mm}
\Paragraph{C.6 Why DCFace is better in  U,D,C metrics?} We note DCFace is not better in all U,D,C. Fig.~6 (main) shows SynFace has the highest consistency (C). But, DCFace excels in the tradeoff between C and D. In other words, style similarity to the real dataset (\textit{i.e.} D) is lacking in other datasets and it is as important as ID consistency. As such, U,D,C metrics reveal weak/strong points of synthetic datasets.  

\clearpage
\section{Visualizations}

\subsection{Time-step Visualizaton}

\begin{figure}[h]
    \centering
    \includegraphics[width=\linewidth]{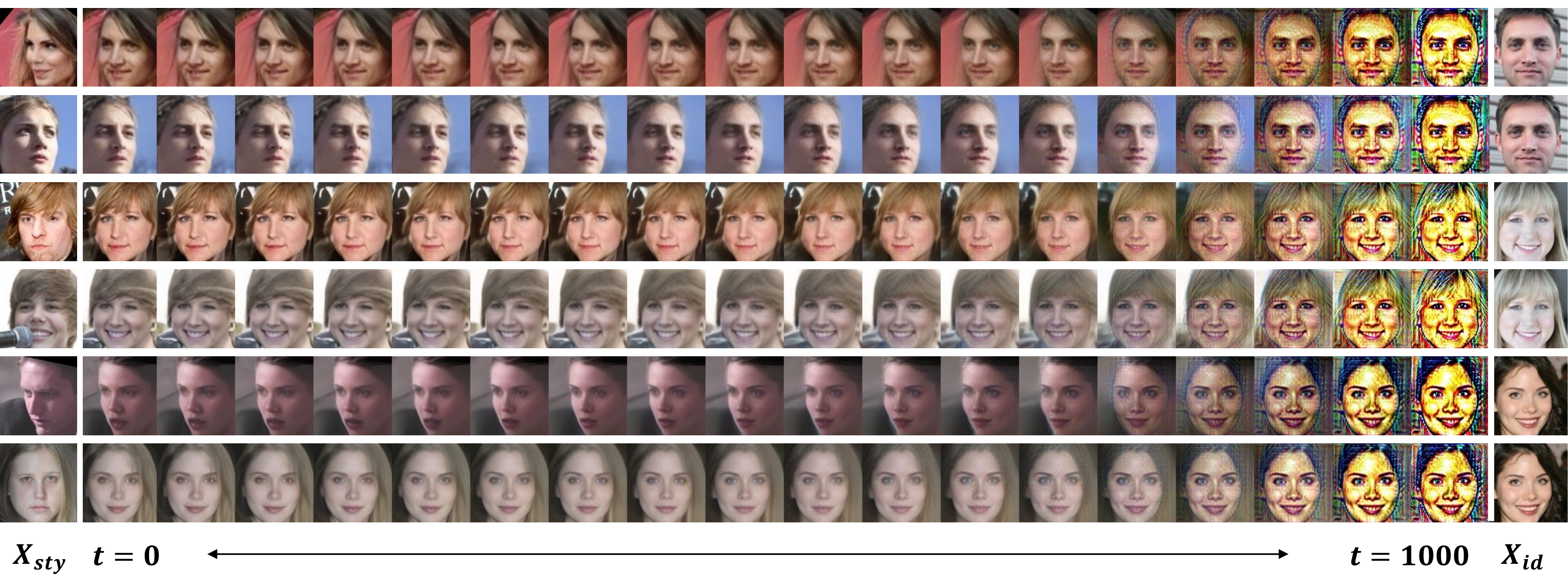}
    \caption{A plot of DCFace outputs at each time-step. }
    \label{fig:timestep}
\end{figure}

Fig.~\ref{fig:timestep} shows how DDPM generates output at each time-step. The far left column shows $\bm{X}_{sty}$, the desired style of an image. The far right column shows $\bm{X}_{id}$, the desired ID image of choice. In early time-steps, the network reconstructs the front-view image with an ID of $\bm{X}_{id}$. And gradually, it interpolates the image into the desired style of $\bm{X}_{sty}$. The gradual transition can be in the pose, hair-style, expression, etc. 

\subsection{Interpolation}
\begin{figure}[h]
    \centering
    \includegraphics[width=\linewidth]{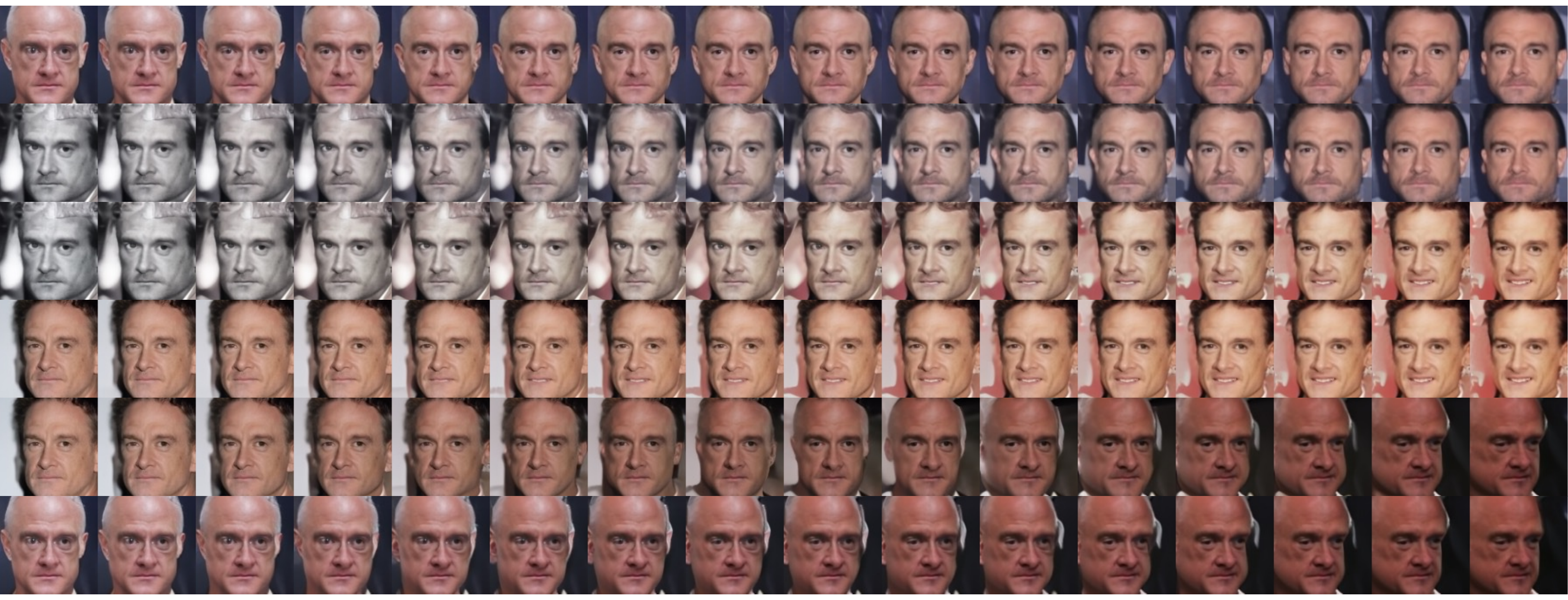}
    \caption{A plot of DCFace output with style interpolation.}
    \label{fig:interpolate}
\end{figure}

In Fig.~\ref{fig:interpolate}, we show the plot of interpolation in $\bm{X}_{sty}$. While keeping the same identity $\bm{X}_{id}$, we take two style images $\bm{X}_{sty1}$ and $\bm{X}_{sty2}$. We interpolate with $\alpha$ in $\alpha E_{stry}(\bm{X}_{sty1}) + (1-\alpha) E_{stry}(\bm{X}_{sty2})$ with $\alpha$ increasing linearly from $0$ to $1$. The interpolation is smooth, creating an intermediate pose and expression that did not exist before. 

\section{Miscelaneous}

\Paragraph{Similarity threshold}
Threshold=0.3 is based on FR evaluation model having a threshold of $0.3080$ for verification with TPR@FPR=$0.01\%:97.17$\% on IJB-B~\cite{ijbb}. FPR=$0.01\%$ is widely used in practice and the scale of similarity is $(\!-\!1\!,1\!)$. 
At threshold=0.3, FFHQ has 200 (2\%) more unique subjects than DDPM, signaling a similar level of uniqueness. 

\Paragraph{Style Extracting Model} We use the early layers of face recognition model for style extractor backbone. Our rationale for adopting the early layers of the FR model, as opposed to that of the ImageNet-trained model is that the early layers extract low-level features and we wanted features optimized with the face dataset. But, it is possible to take other models as long as it generates low-level features.

\Paragraph{Evaluation on Harder Datasets} We evaluate on harder datasets,
IJB-B~\cite{ijbb} (TPR@FPR=0.01\%:~$75.12$) and TinyFace~\cite{tinyface} (Rank1:~$41.66$). We include this result for future works to evaluate on harder datasets.

\Paragraph{Real and Generated Similarity Analysis}
 In addition to Fig.7 mathcing $\hat{X}_{id}$ with CASIA-WebFace, matching all $\hat{X_0}$ (generated) images against CASIA-WebFace at threshold=0.3, we get 0.0026\% FMR. This implies that only a small fraction of CAISA-WebFace images are similar to the generated images.   

\section{Societal Concerns}

We believe that the Machine Learning and Computer Vision community should strive together to minimize the negative societal impact. Our work falls into the category of 1) image generation using generative models and 2) synthetic labeled dataset generation. In the field of image generation, unfortunately, there are numerous well-known malicious applications of generative models. Fake images can be used to impersonate high-profile figures and create fake news. Conditional image generation models make the malicious use cases easier to adapt to different use cases because of user controllability. Fortunately, GAN-based generators produce subtle artifacts in the generated samples that allow the visual forgery 
detection~\cite{wang2020cnn,yu2019attributing,girish2021towards,proactive}. With the recent advance in DDPM, the community is optimistic about detecting forgeries in diffusion models~\cite{sha2022fake}. It is also known that proactive treatments on generated images increase the forgery detection performance~\cite{proactive}, and as generative models become more sophisticated, proactive measures may be advised whenever possible.  
 
Synthetic dataset generation is, on the other hand, an effort to avoid infringing the privacy of individuals on the web. Large-scale face dataset is collected without informed consent and only a few evaluation datasets such as IJB-S~\cite{ijbs} has IRB compliance for safe and ethical research. Collecting large-scale datasets with informed consent is prohibitively challenging and the community uses web-crawled datasets for the lack of an alternative option. Therefore, efforts to create synthetic datasets with synthetic subjects can be a practical solution to this problem. In our method, we still use real images to train the generative models. We hope that research in synthetic dataset generation will eventually replace real images, not just in the recognition task, but also in the generative tasks as well, removing the need for using real datasets in any form.   

\section{Implementation Details and Code}
The code will be released at \url{https://github.com/mk-minchul/dcface}. 
For preprocessing the training data CASIA-WebFace~\cite{casia}, we reference AdaFace~\cite{kim2022adaface} and use  MTCNN~\cite{zhang2016joint} for alignment and cropping faces. For the backbone model definition, TFace~\cite{TFace} and for evaluation of LFW~\cite{lfw}, CFP-FP~\cite{cfpfp}, CPLFW~\cite{cplfw}, AgeDB~\cite{agedb} and CALFW~\cite{calfw}, we use AdaFace repository ~\cite{kim2022adaface}.

\end{document}